%% file: main.tex
\documentclass[10pt,twocolumn,letterpaper]{article}

\usepackage{cvpr}
\usepackage{times}
\usepackage{epsfig}
\usepackage{graphicx}
\usepackage{amsmath}
\usepackage{amssymb}
\usepackage{multirow}
\usepackage{multicol}
\usepackage{booktabs} 
\usepackage[export]{adjustbox} 
\usepackage{longtable}
\usepackage{collcell}
\usepackage{mathrsfs,amsmath,mathabx} 
\usepackage{adjustbox}
\usepackage{graphicx}
\usepackage{pifont} % for check/xmark
\usepackage{xcolor,colortbl} % for highlihgting columns
\usepackage{arydshln} % dashed lines
\usepackage{algorithm}
\usepackage{algorithmic} % pseudocode
\usepackage{textcomp} % quote
\usepackage{placeins}
\usepackage{tabularx}

\input{macros.tex}

\newcommand{\cmark}{\ding{51}}%
\newcommand{\xmark}{\ding{55}}%

\definecolor{Gray}{gray}{0.85}
\newcolumntype{g}{>{\columncolor{Gray}}c}

\def\eatcell#1\unskip{}
\newcolumntype{E}{>{\eatcell}c@{}}

\usepackage[pagebackref=true,breaklinks=true,letterpaper=true,colorlinks,bookmarks=false]{hyperref}

\cvprfinalcopy % *** Uncomment this line for the final submissio

\ifcvprfinal\fi
\begin{document}

%%%%%%%%% TITLE
\title{15 Keypoints Is All You Need}

\author{
    Michael Snower$^\dagger$\thanks{Work done as a NEC Labs intern}\quad Asim Kadav$^\ddagger$\quad Farley Lai$^\ddagger$\quad Hans Peter Graf$^\ddagger$ \\
    $^\dagger$Brown University \quad $^\ddagger$NEC Labs America \\
    {\tt\small michael\_snower@brown.edu \quad \{asim,farleylai,hpg\}@nec-labs.com}
}

\maketitle

\input{sections/abstract}
\input{sections/intro}

\input{sections/related}

\input{sections/methods}

\input{sections/results}

\input{sections/analysis}
\input{sections/conclusion}

{\small
\bibliographystyle{ieee_fullname}
\bibliography{egbib}
}

\newpage\phantom{blabla}
\newpage\phantom{blabla}
\appendix
\vspace{-0.75cm}
\section{Supplementary Material for \trackname}

\subsection{Test Set Scores} We submitted to the PoseTrack 2017 test set twice. We first achieved a \textbf{60.1} MOTA score, but then decreased the TOKS box expansion value from $\alpha=1.4$ to $\alpha=1.25$. This increased our our score to \textbf{61.2}. $\alpha = 1.25$ is also the value we used on the 2018 Validation Set.

\subsection{Additional Qualitative Results} 
\begin{figure*}[t!]
    \centering
    \begin{adjustbox}{width=1.0\textwidth}
        \includegraphics{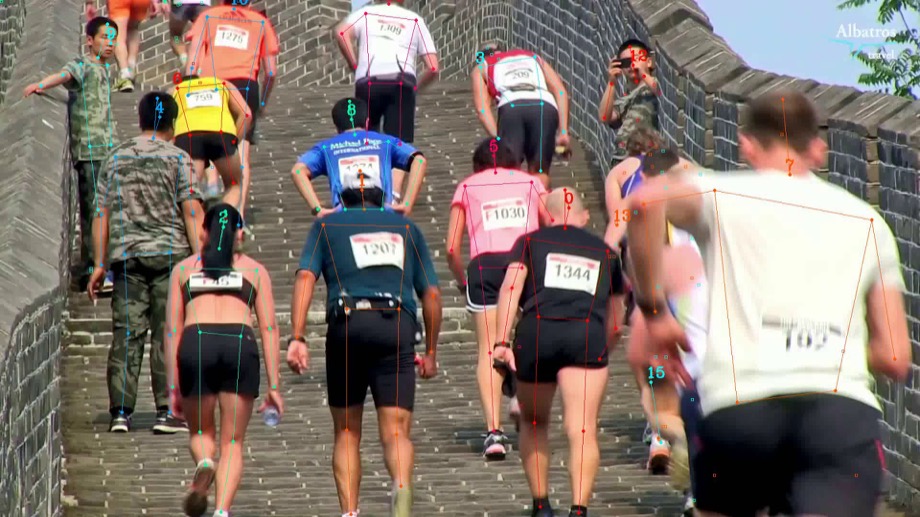}
        \includegraphics{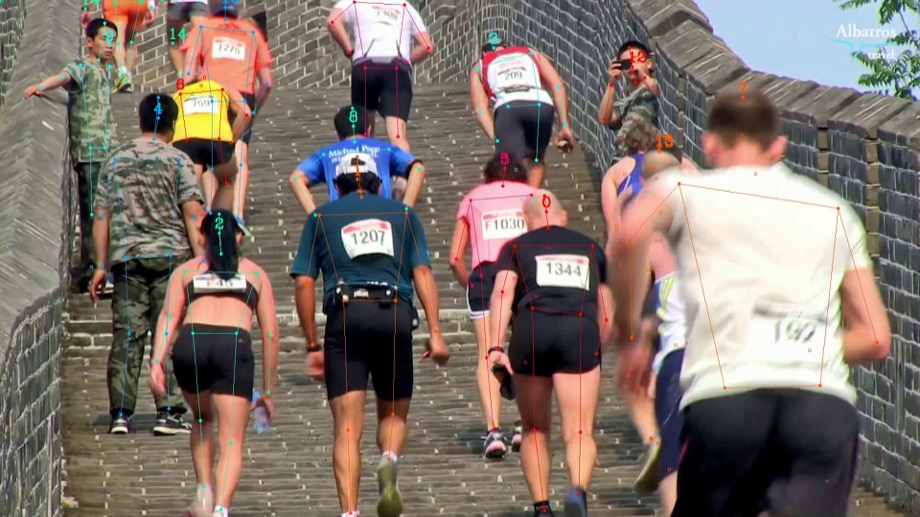}
        \includegraphics{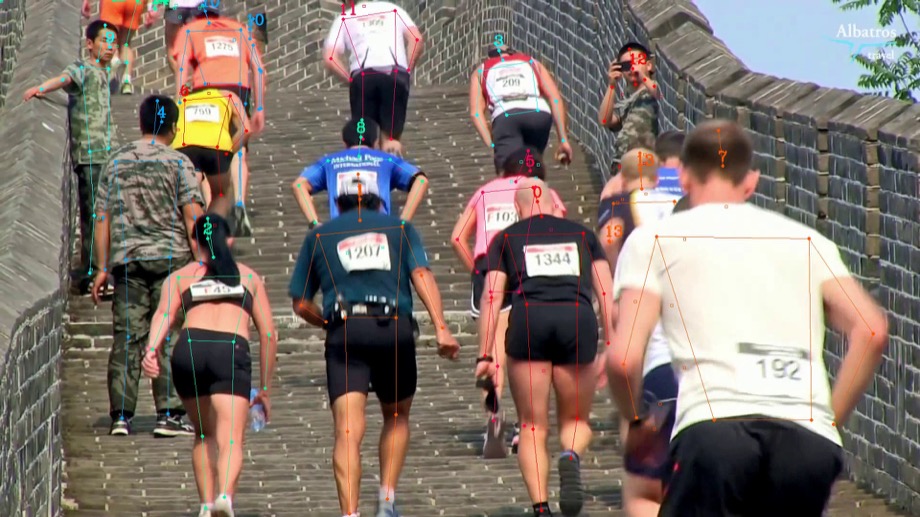}
        \includegraphics{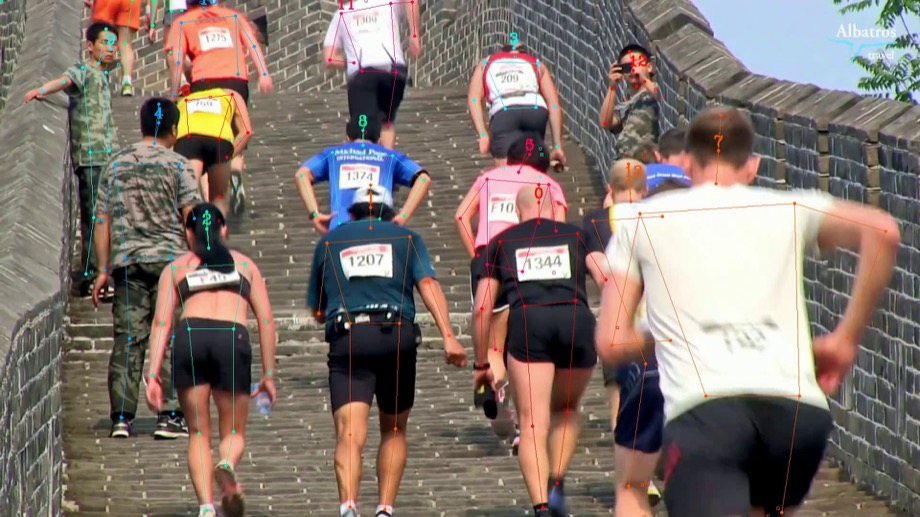}
    \end{adjustbox}
    \tiny
    \begin{tabularx}{\textwidth}{X}
        \phantom{blabla} \\
    \end{tabularx}{}
    \begin{adjustbox}{width=1.0\textwidth}
        \includegraphics{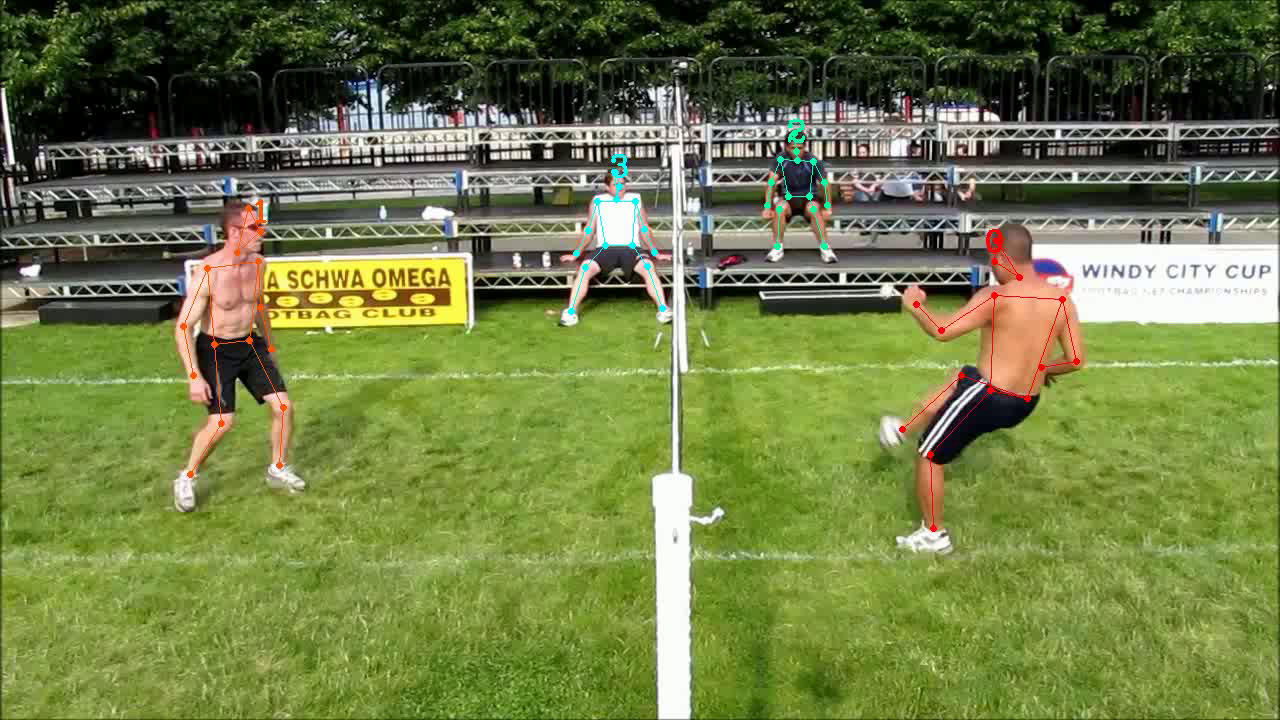}
        \includegraphics{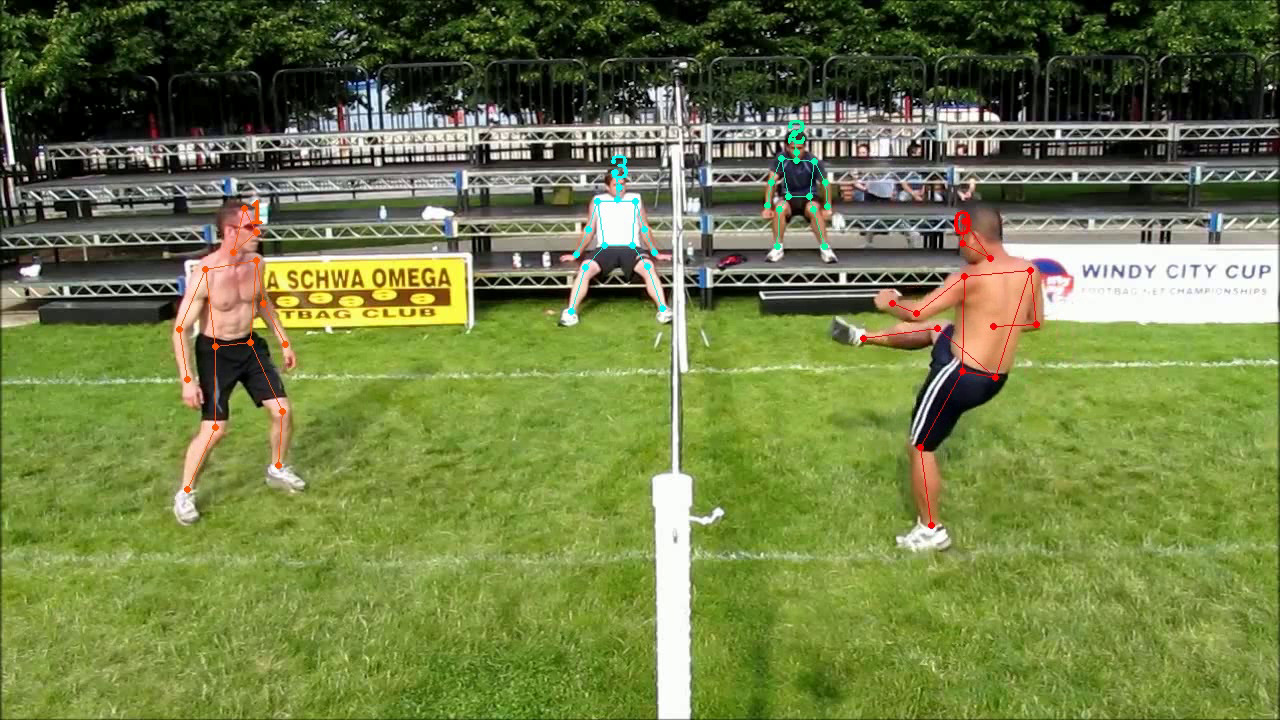}
        \includegraphics{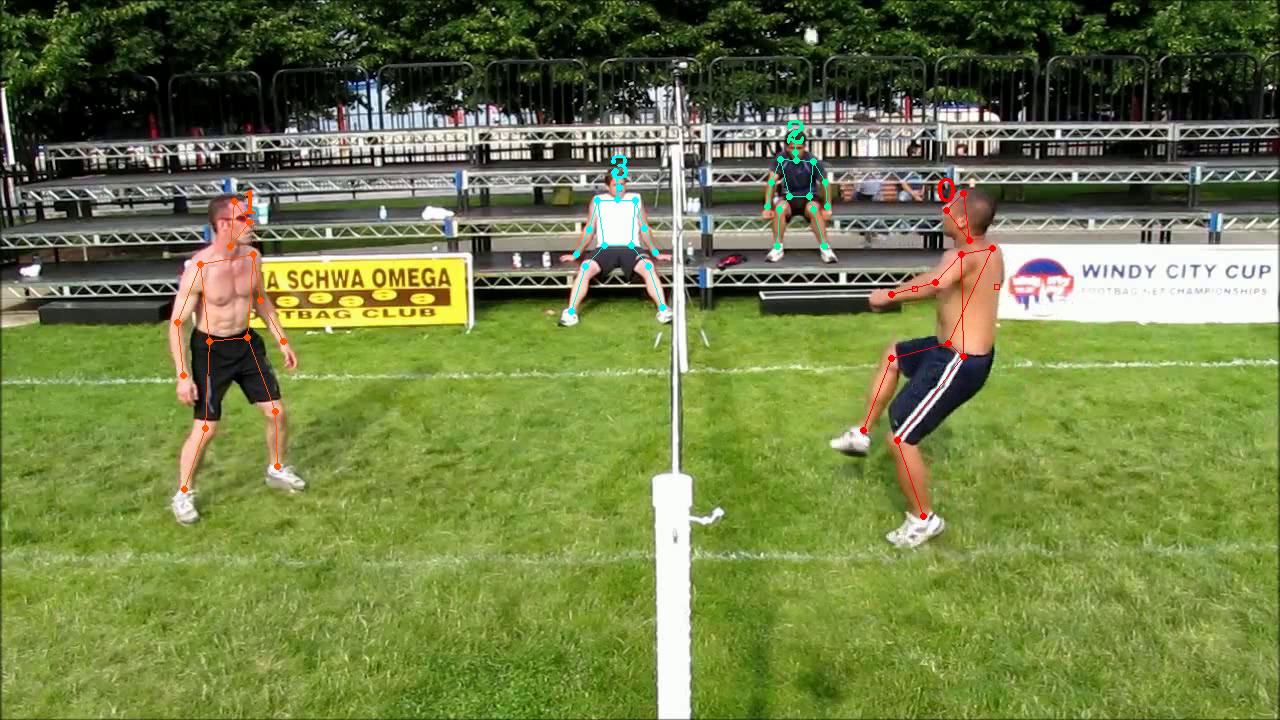}
        \includegraphics{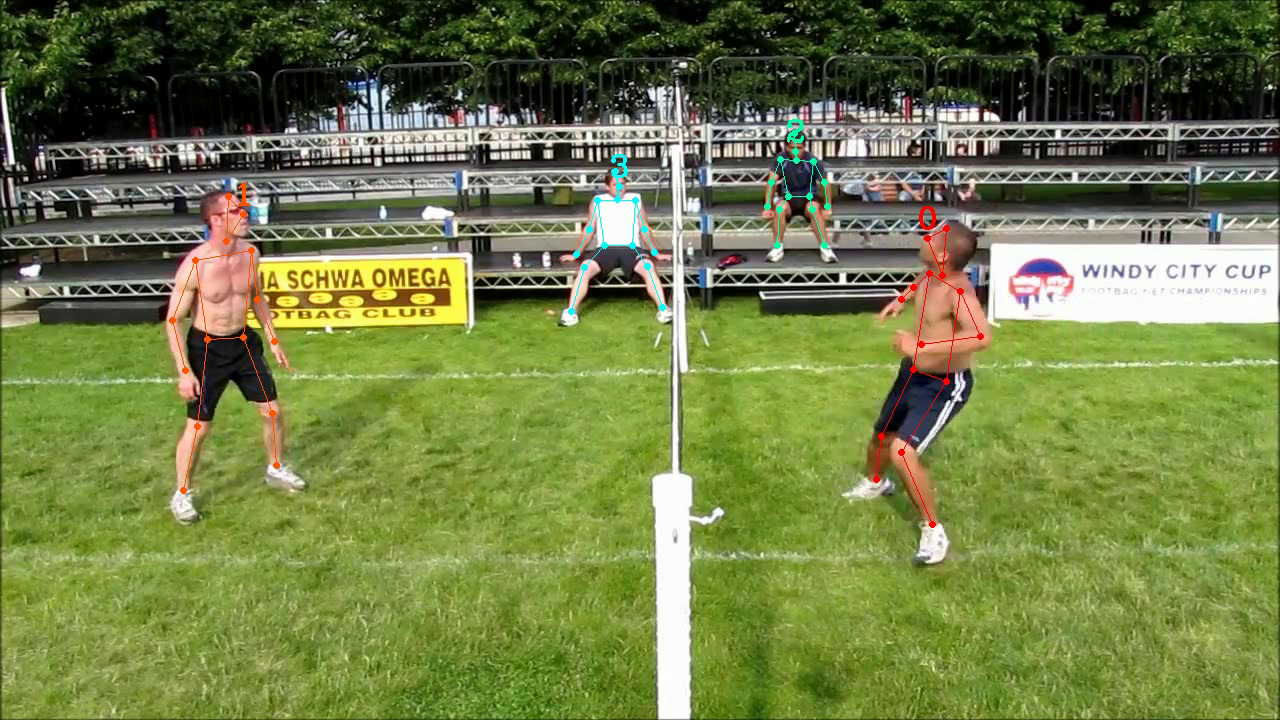}
    \end{adjustbox}
    \begin{tabularx}{\textwidth}{X}
        \phantom{blabla} \\
    \end{tabularx}{}
    \begin{adjustbox}{width=1.0\textwidth}
        \includegraphics{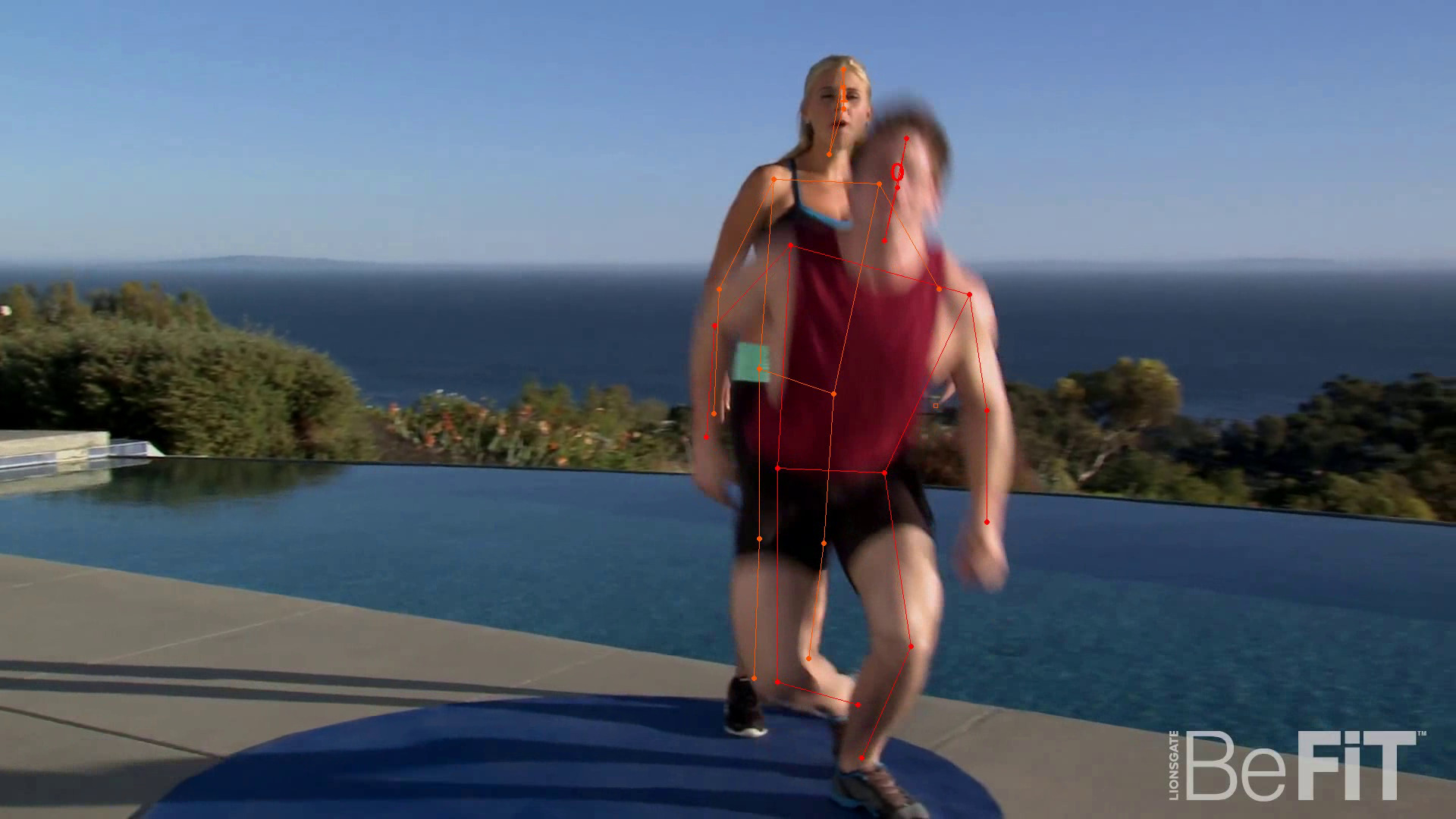}
        \includegraphics{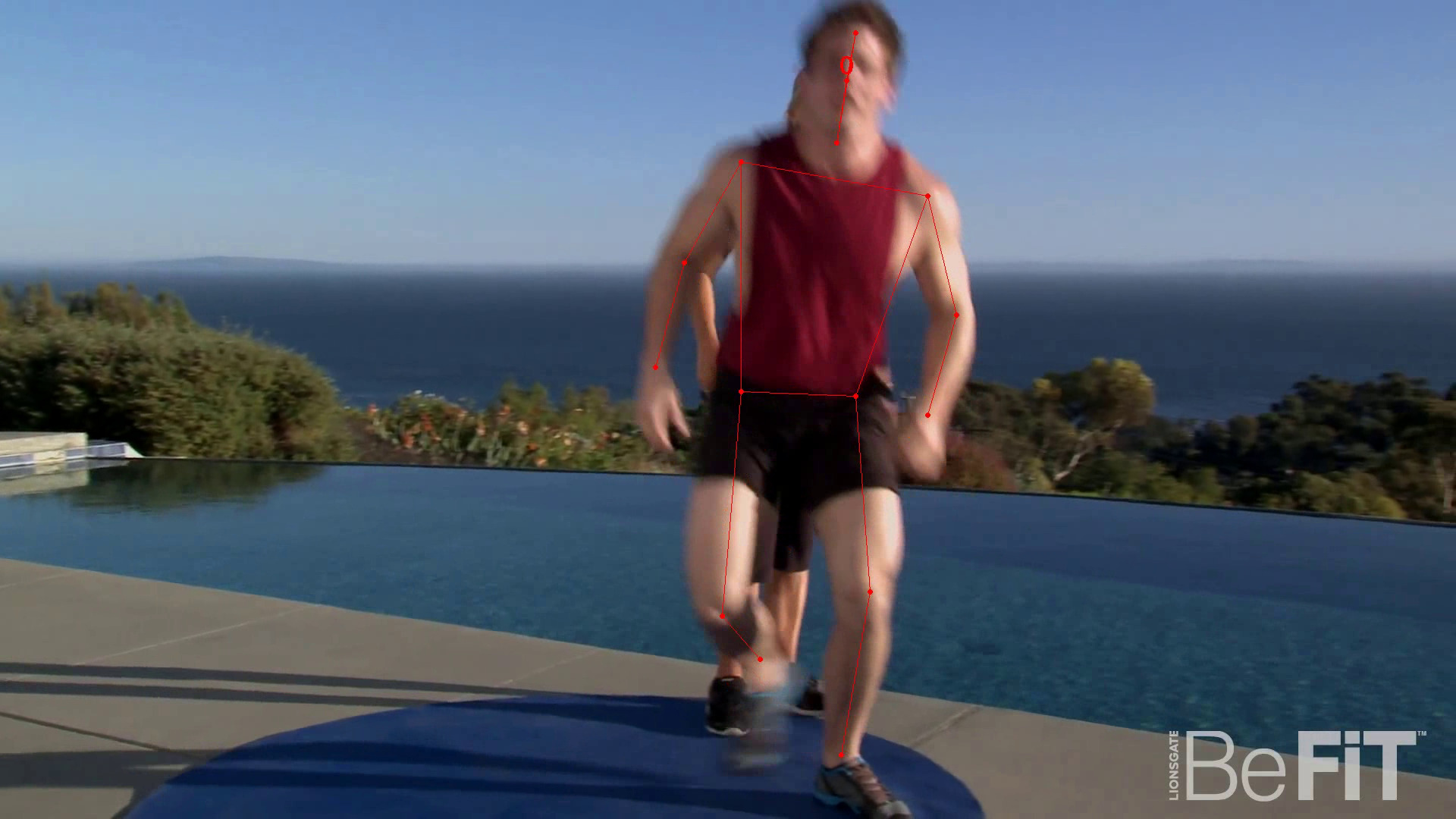}
        \includegraphics{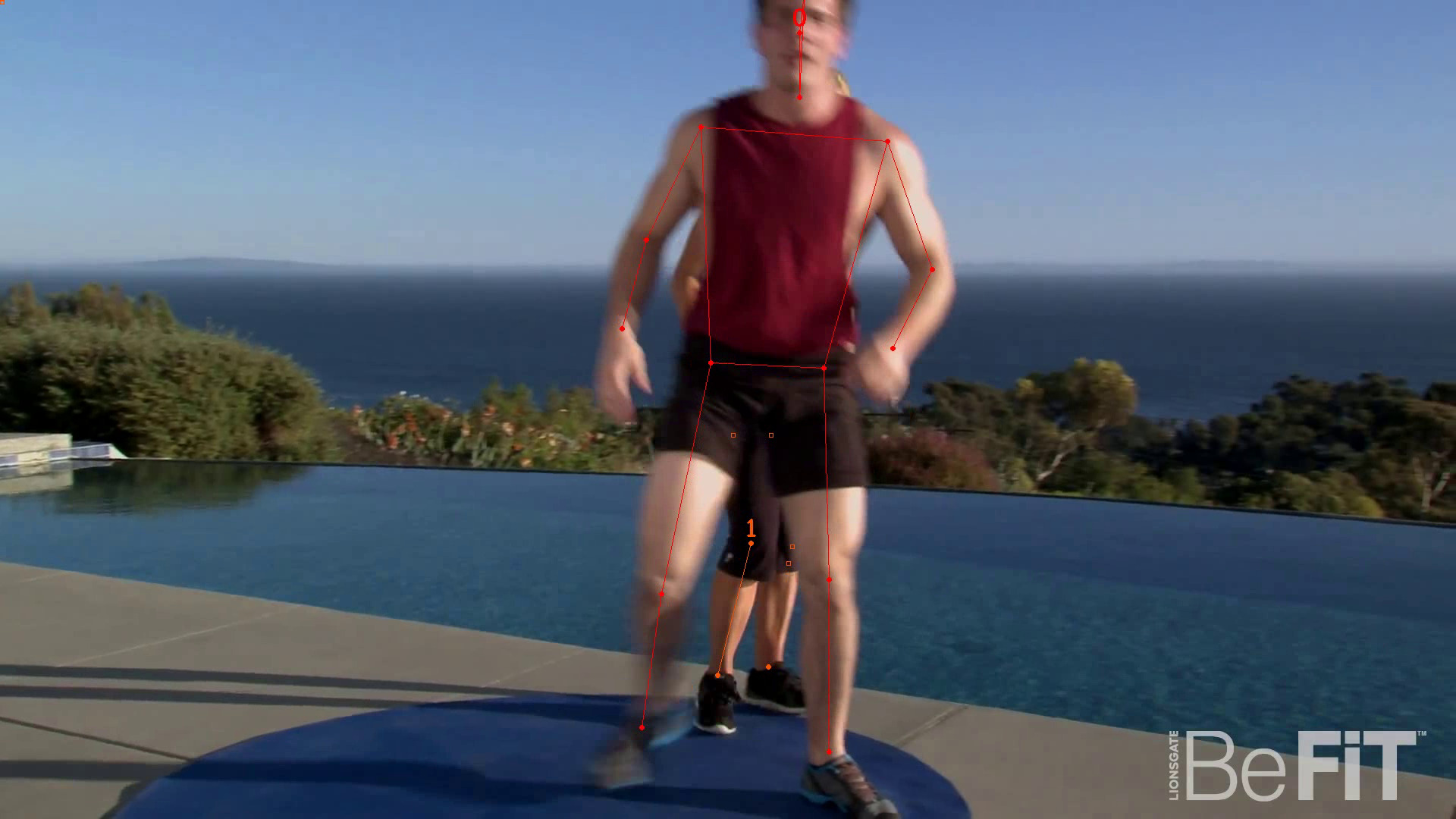}
        \includegraphics{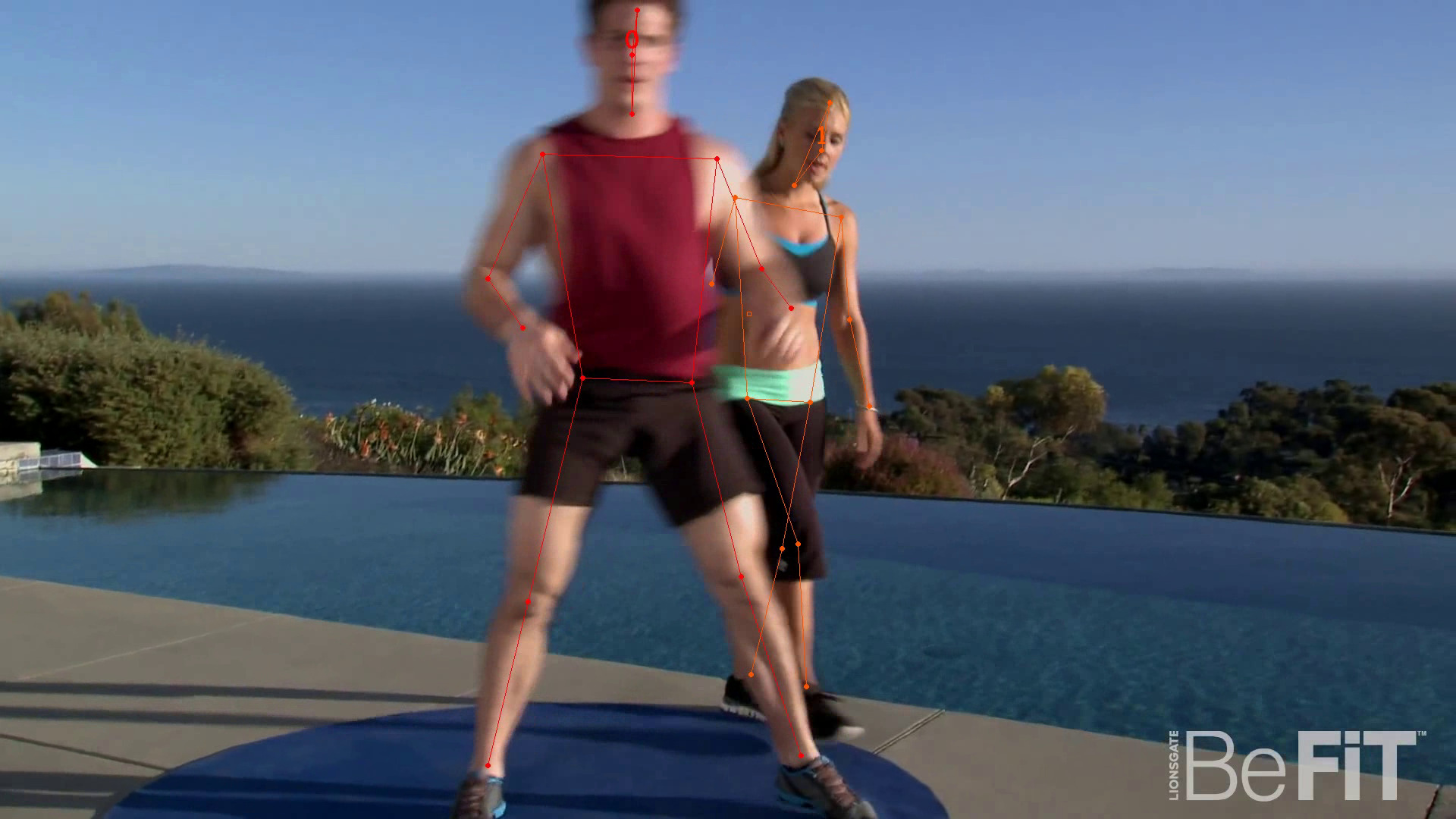}
    \end{adjustbox}
    \begin{tabularx}{\textwidth}{X}
        \phantom{blabla} \\
    \end{tabularx}{}
    \begin{adjustbox}{width=1.0\textwidth}
        \includegraphics{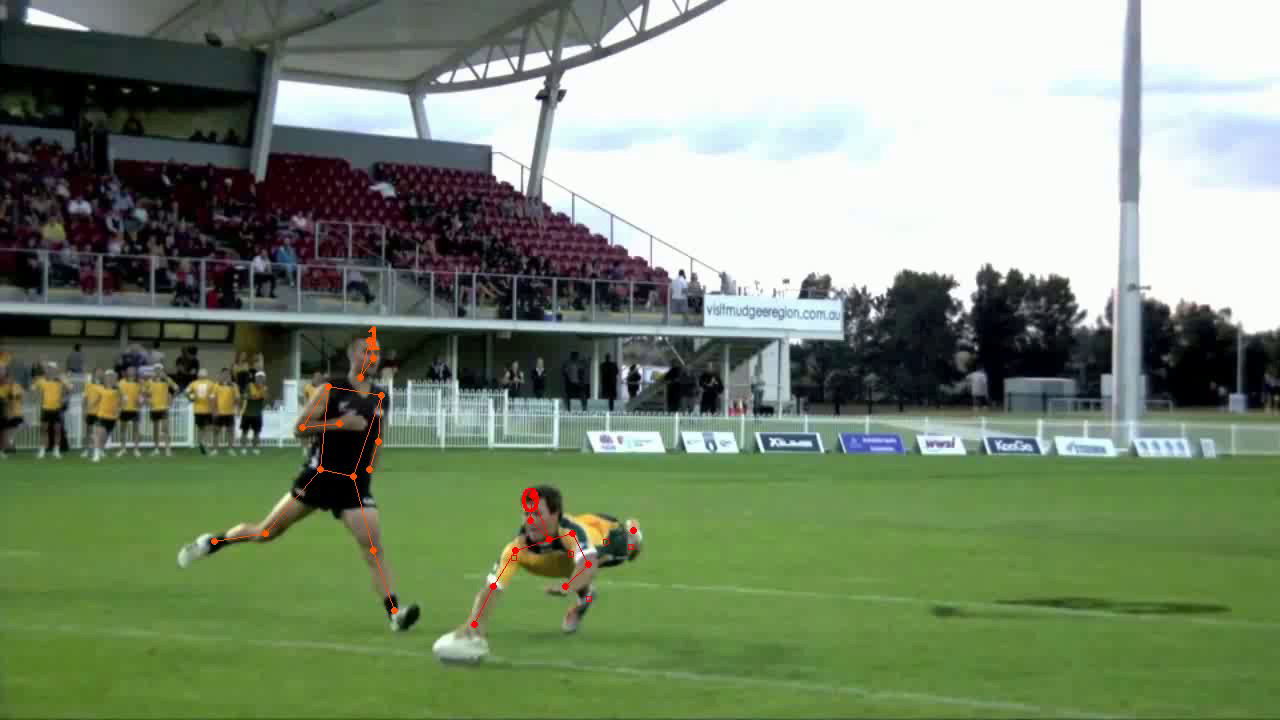}
        \includegraphics{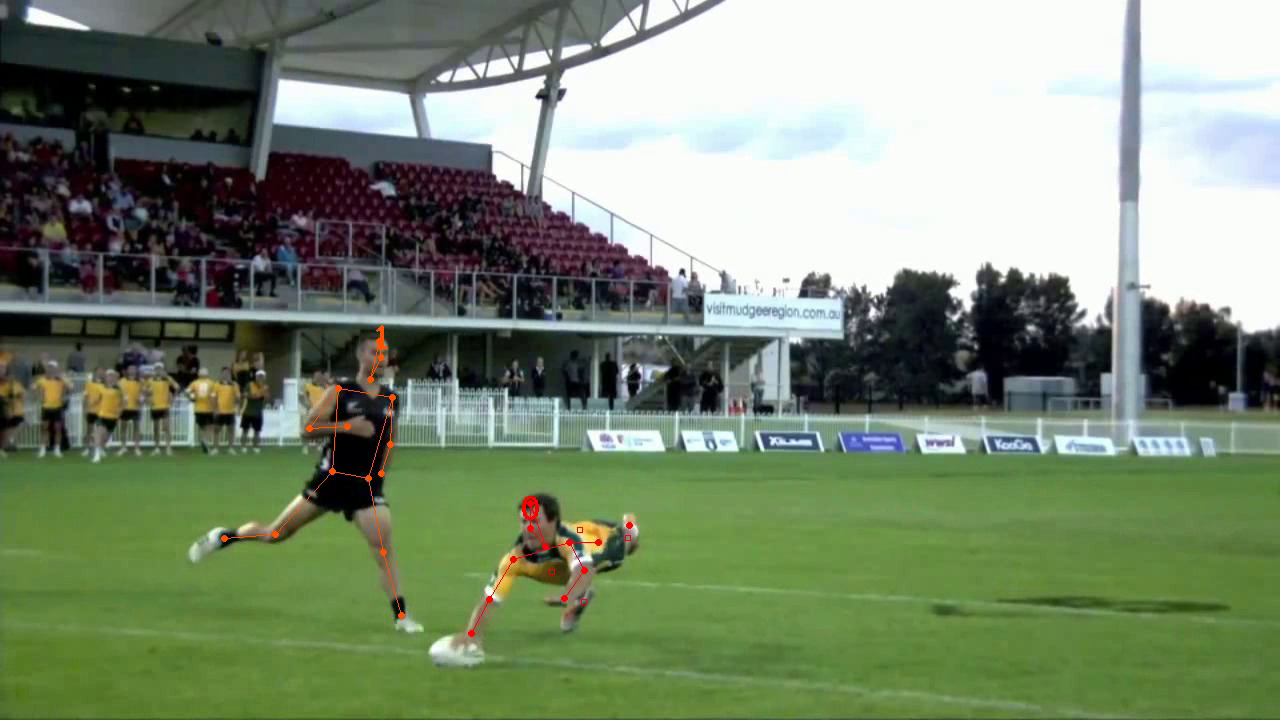}
        \includegraphics{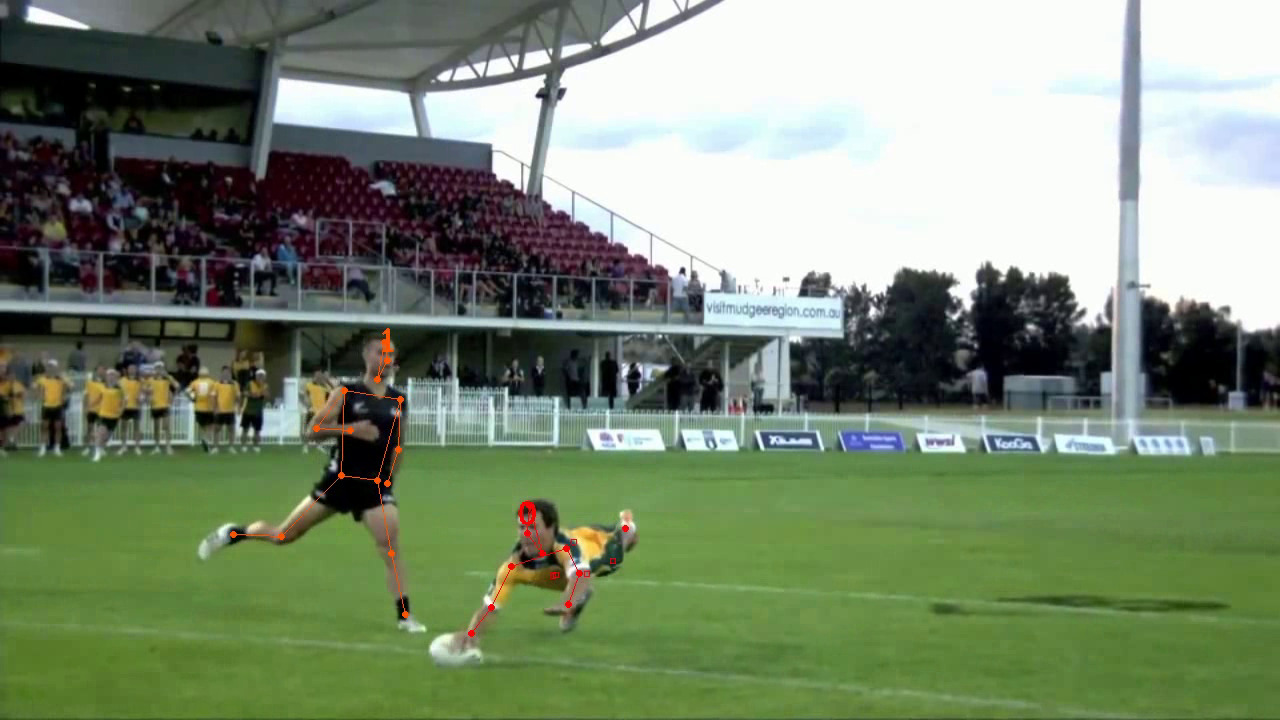}
        \includegraphics{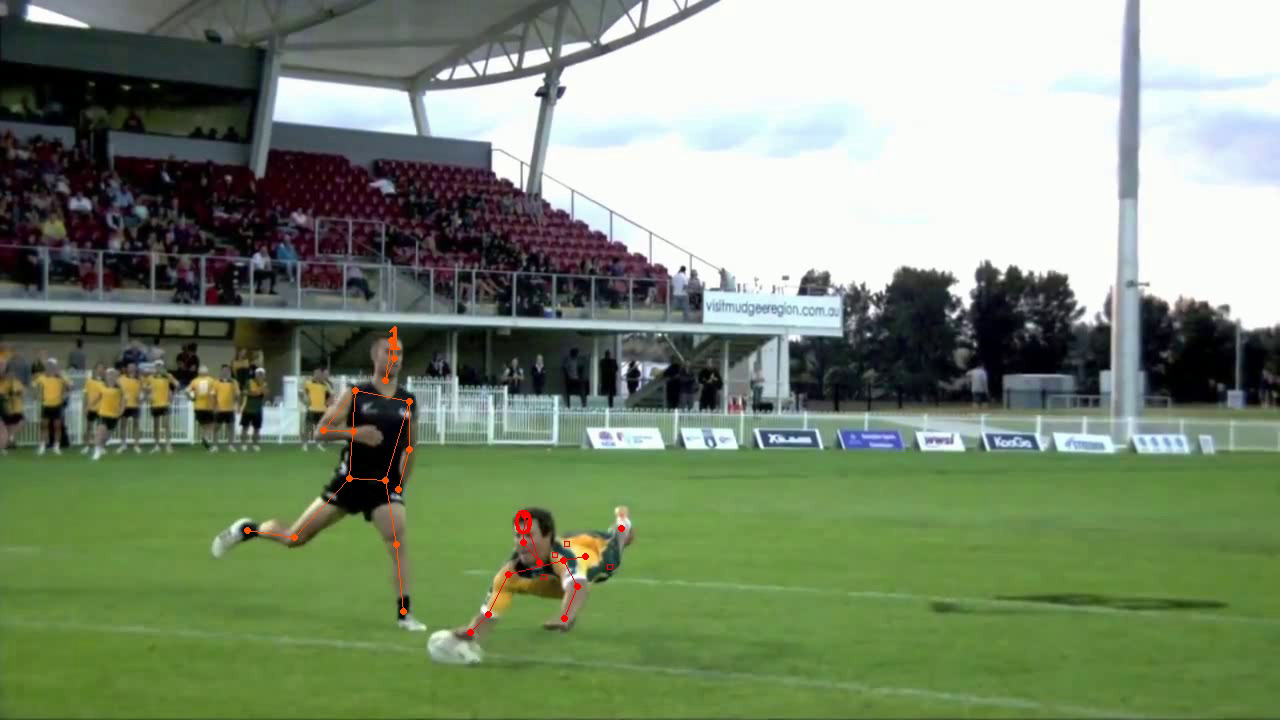}
    \end{adjustbox}
    \begin{tabularx}{\textwidth}{X}
        \phantom{blabla} \\
    \end{tabularx}{}
    \begin{adjustbox}{width=1.0\textwidth}
        \includegraphics{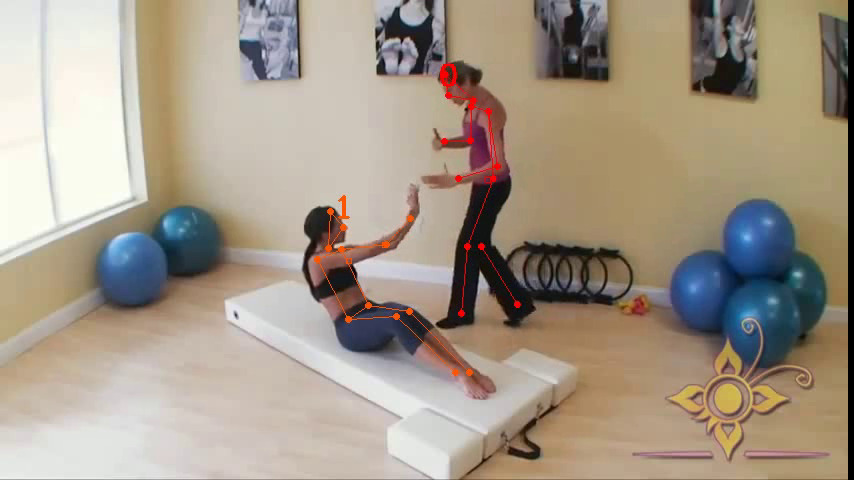}
        \includegraphics{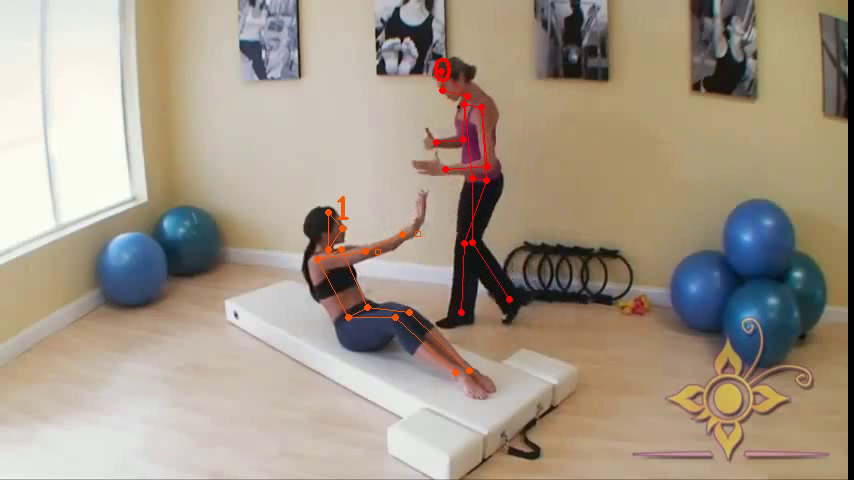}
        \includegraphics{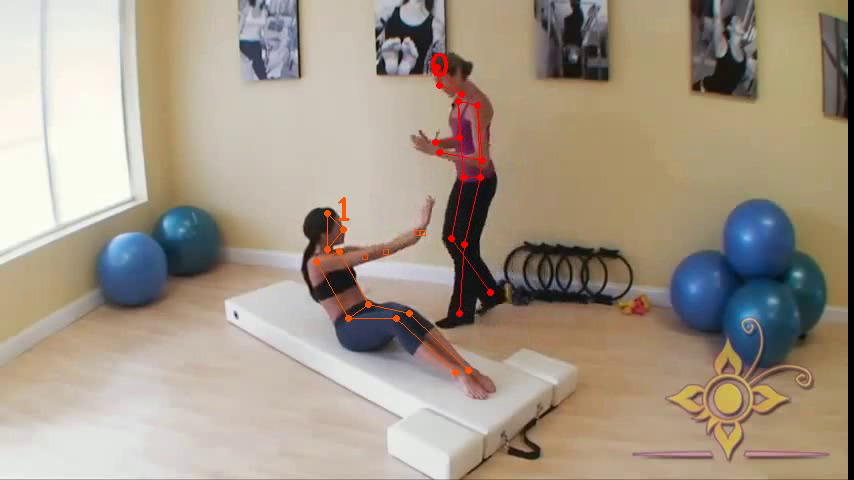}
        \includegraphics{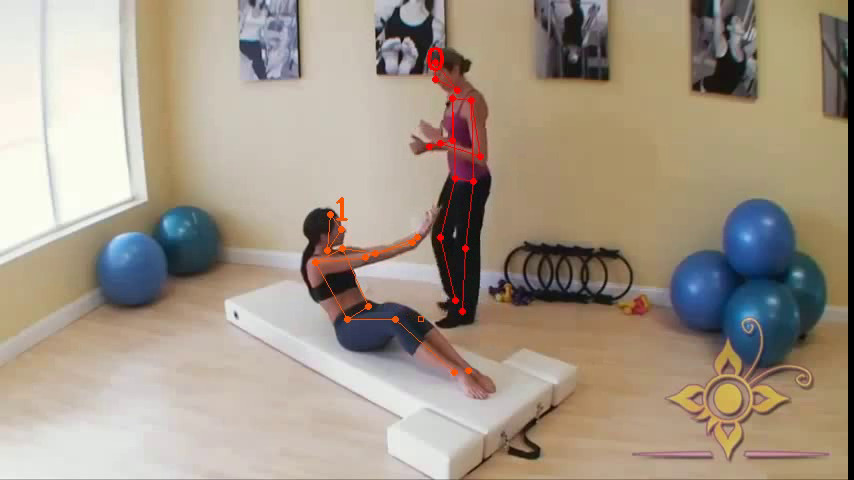}
    \end{adjustbox}
    \begin{tabularx}{\textwidth}{X}
        \phantom{blabla} \\
    \end{tabularx}{}
    \begin{adjustbox}{width=1.0\textwidth}
        \includegraphics{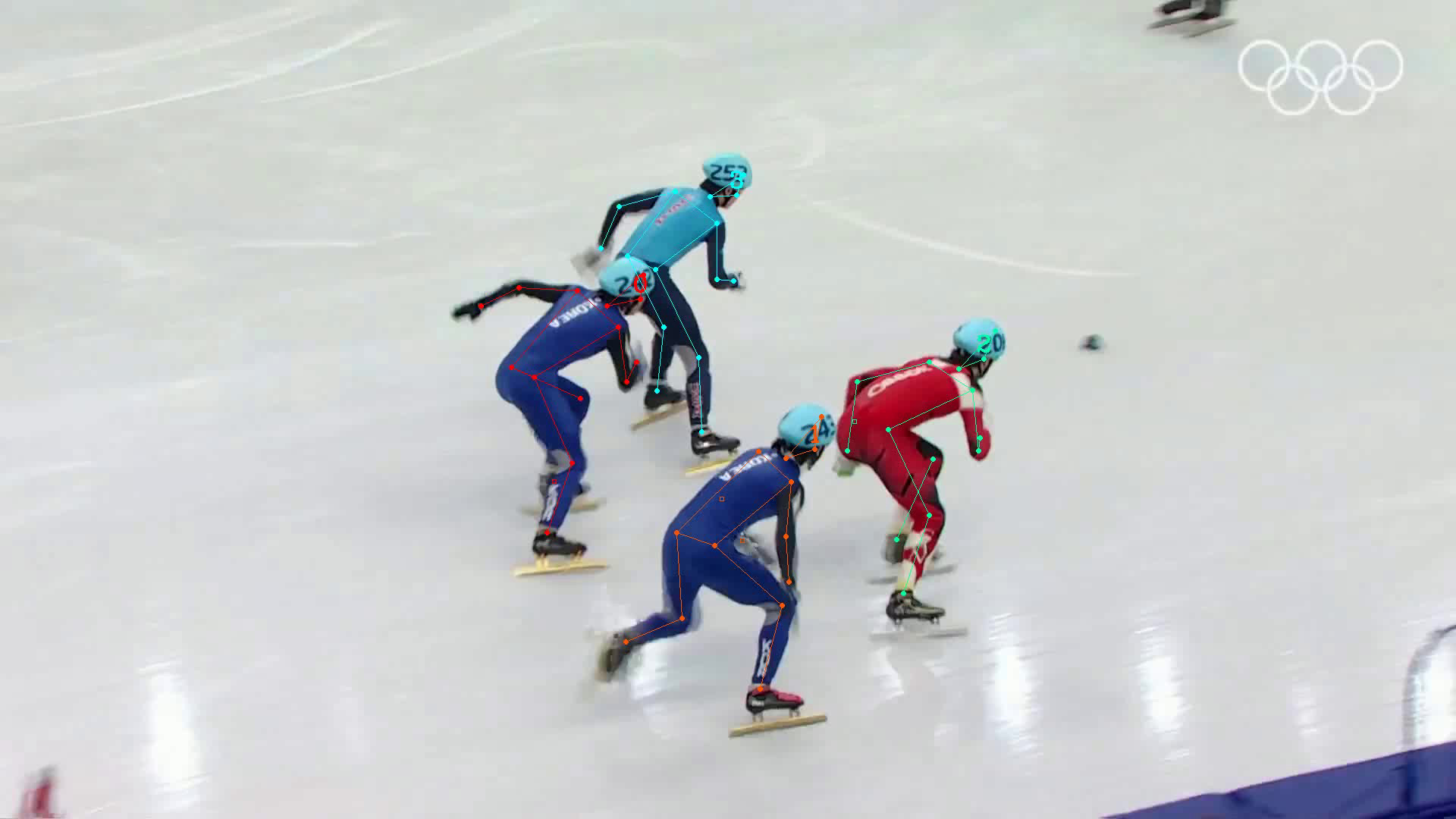}
        \includegraphics{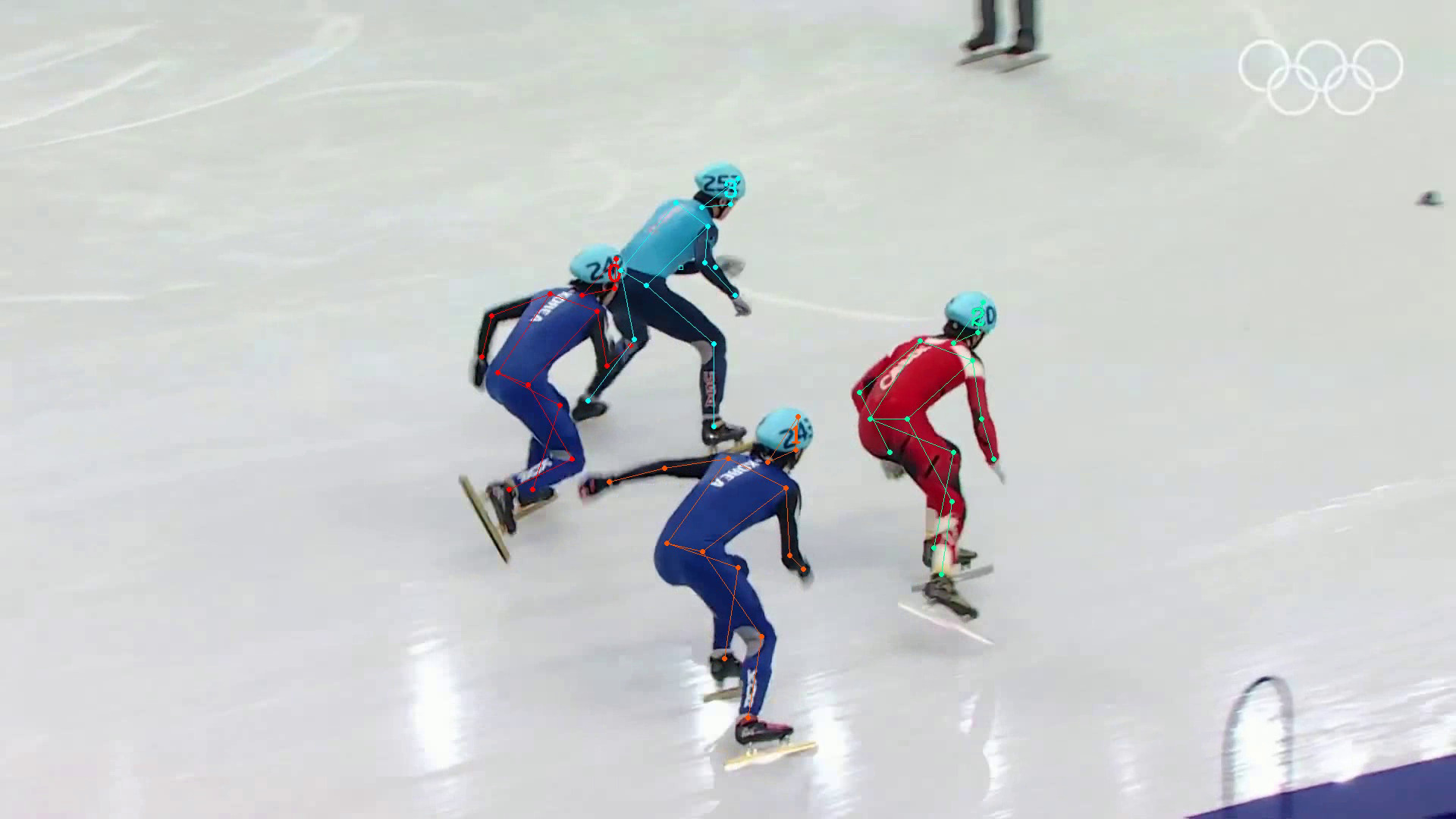}
        \includegraphics{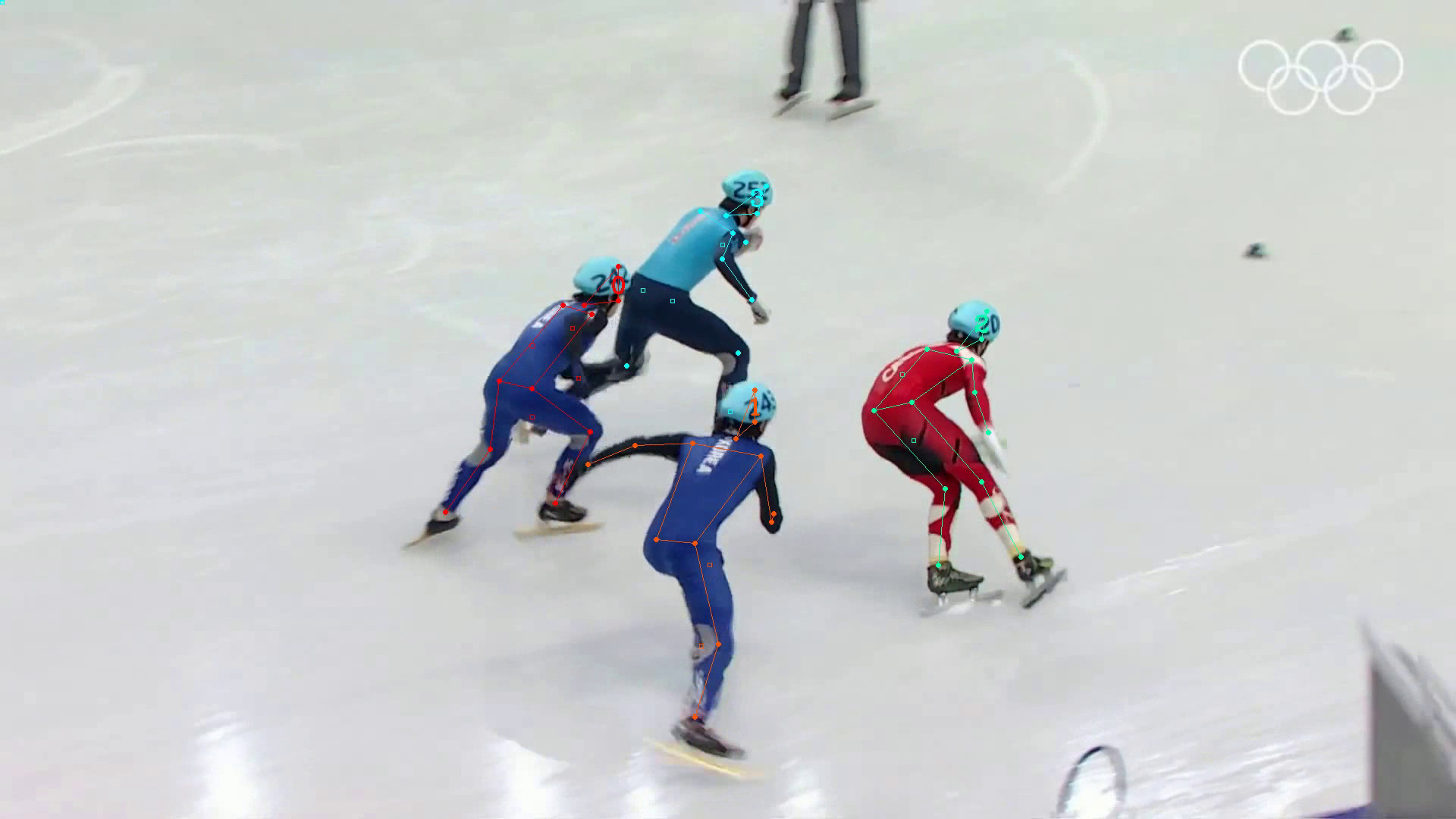}
        \includegraphics{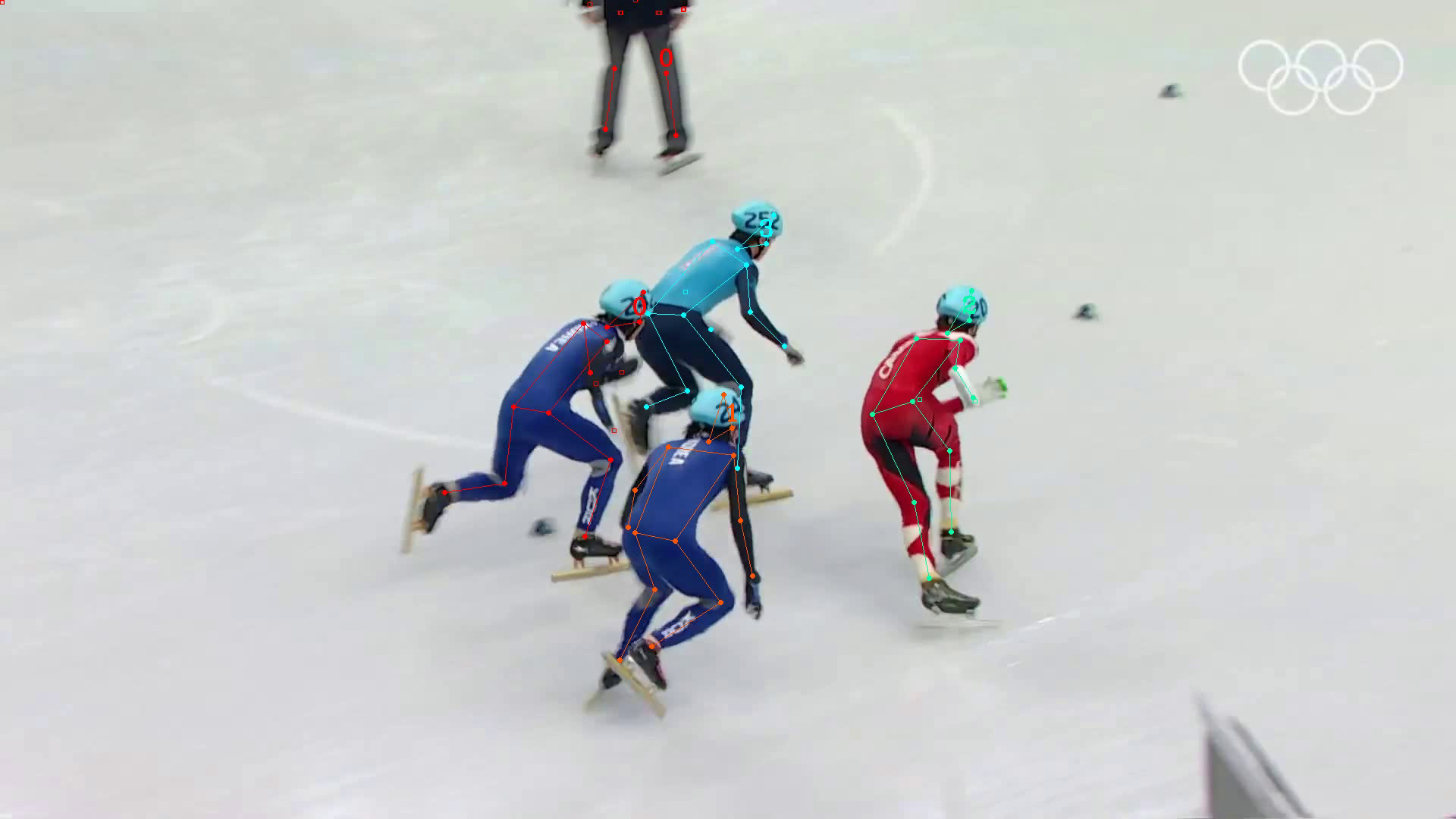}
    \end{adjustbox}
    \begin{tabularx}{\textwidth}{X}
        \phantom{blabla} \\
    \end{tabularx}{}
    \begin{adjustbox}{width=1.0\textwidth}
        \includegraphics{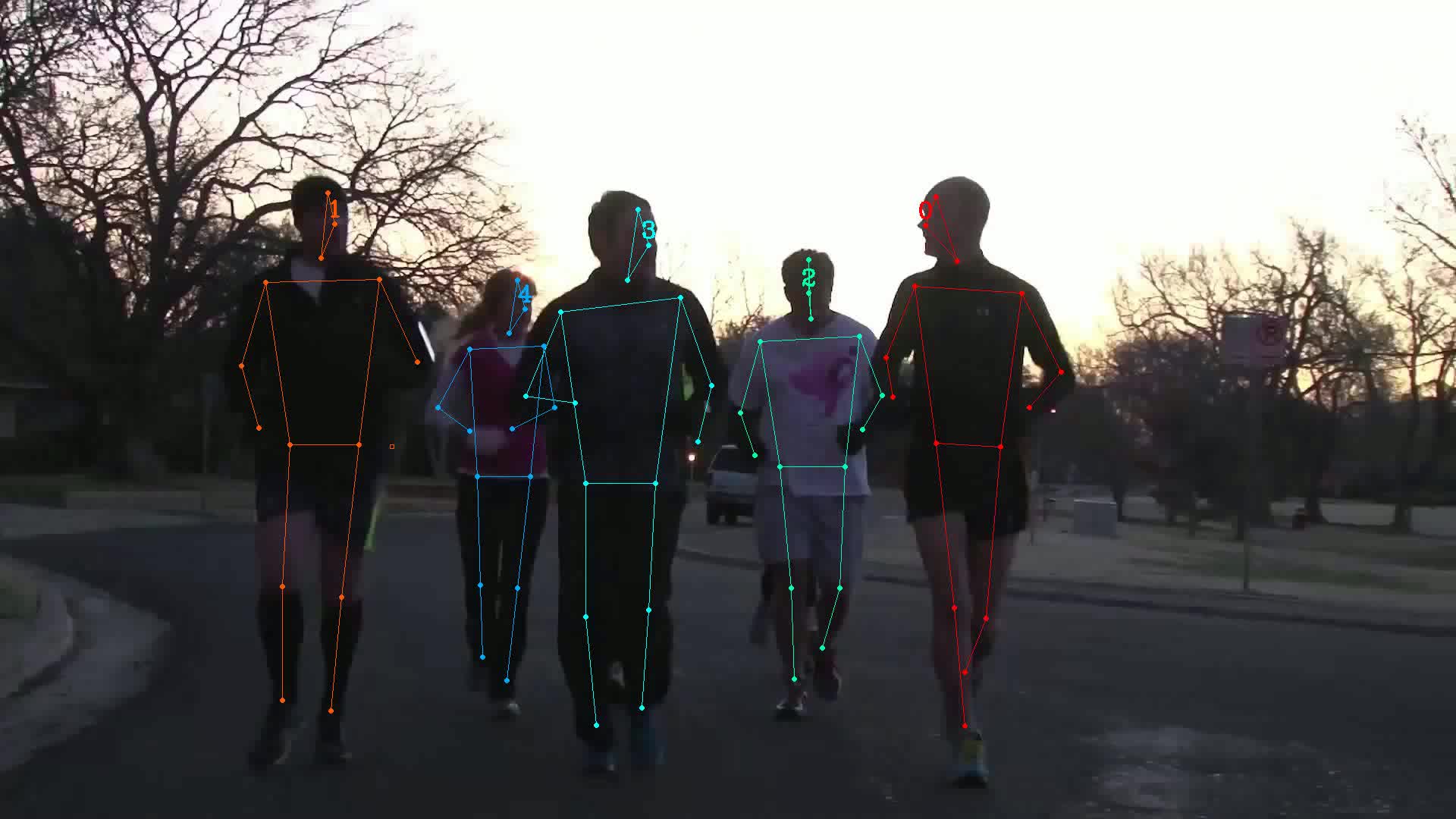}
        \includegraphics{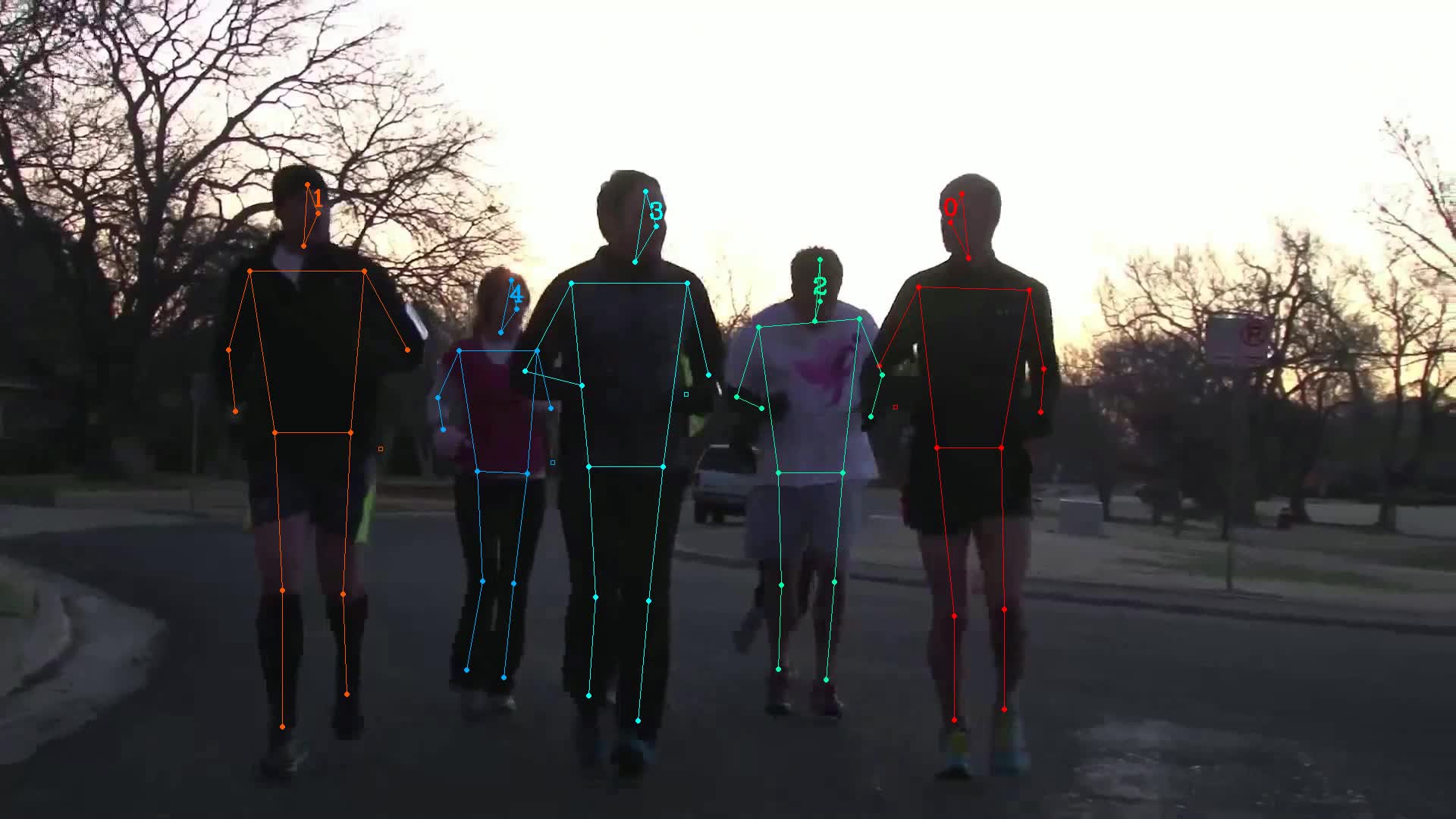}
        \includegraphics{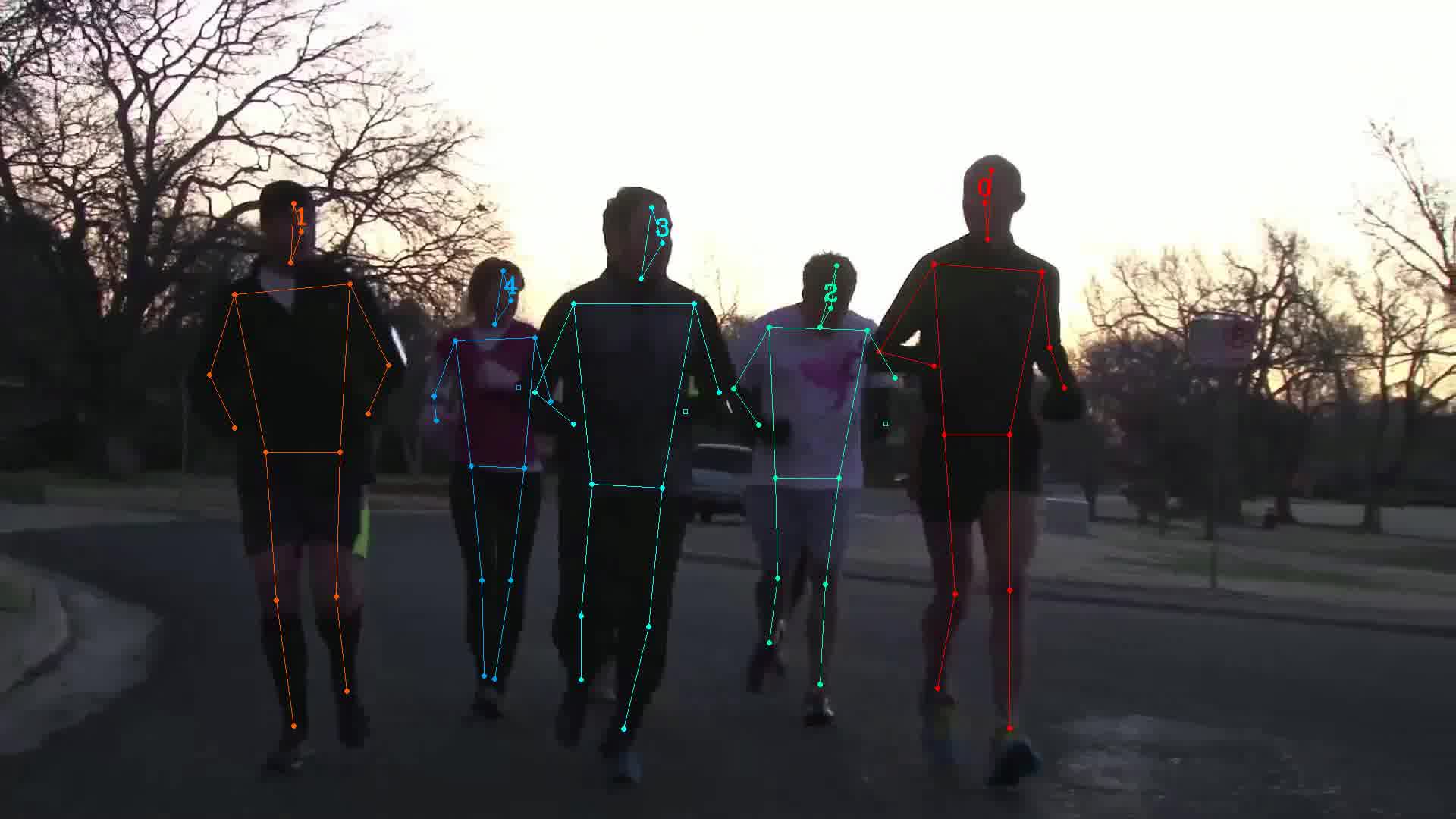}
        \includegraphics{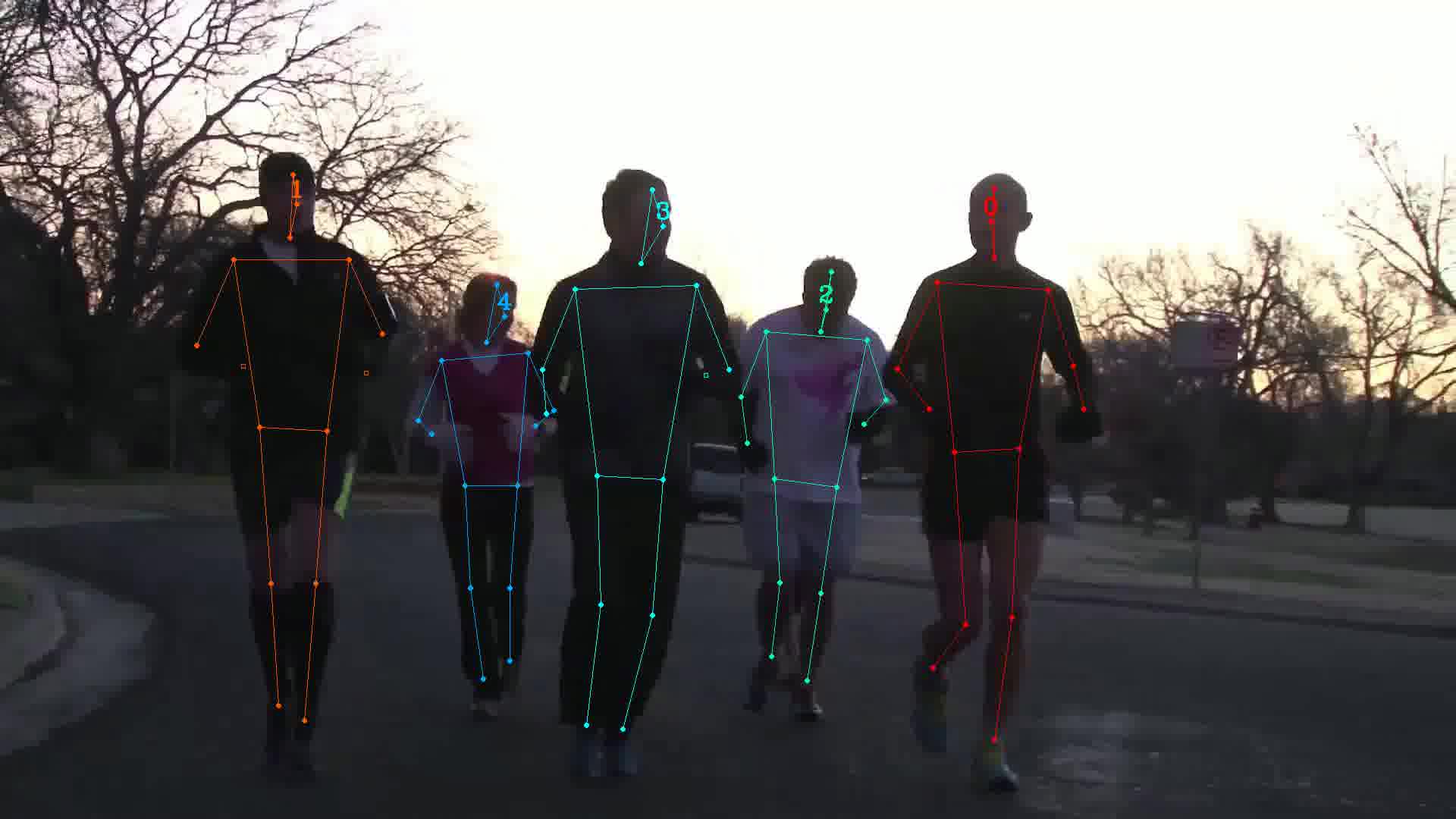}
    \end{adjustbox}
    \caption{Additional qualitative results of our model succeeding in scenarios despite occlusions, unconventional poses, and varied lighting conditions. Every 4th frame is sampled so more extensive motion can be shown. Solid circles represent predicted keypoints. Hollow squares represent keypoint predictions that are not used due to low confidence.}
    \label{fig:additional}
\end{figure*}

We provide additional qualitative results of our model on the PoseTrack 18 Validation Set in Figure \ref{fig:additional}.

\subsection{Keypoint Postprocessing} \label{sec:postprocessing} The post-processing performed when evaluating AP and MOTA is different. Specifically, we use a different keypoint confidence threshold, where keypoints above the threshold are kept and keypoints below the threshold are ignored. The confidence metric used is the per-keypoint confidence score from the pose estimator. The threshold optimal for MOTA is much higher than AP. Interestingly, ID Switches are not much worse, indicating the majority of the error stems from the estimation step. Results are in Table \ref{tab:postprocessing}.

\begin{table}[h]
    \centering
    \begin{tabular}{l | c c c }
        \toprule
        Confidence Threshold & AP & \% IDSW & MOTA \\
        \midrule
        0.05 & \bf 81.6 & 1.0 & 42.0 \\
        0.35 & 79.6 & 0.9 & 63.3 \\
        0.5 & 76.7 & 0.9 & 66.5 \\
        0.57 & 74.3 & \bf 0.8 & \bf 66.6 \\
        0.6 & 72.8 & 0.9 & 66.0 \\
        \bottomrule
    \end{tabular}
    \caption{Effect of postprocessing on the 2018 Validation Set.}
    \label{tab:postprocessing}
\end{table}{}

\subsection{Implementation Details} 

\paragraph{Training} 
To fine-tune the detector, separate models are fine-tuned on PoseTrack 17 and 18 datasets for 1 epoch with a learning rate of $1.9 \times 10^{-3}$ and batch size of 4. Training was conducted on 4 NVIDIA GTX Titans. To fine tune the pose estimator, originally trained on COCO, we follow \cite{HRNet}.

During tracking training, we use a linear warm up schedule for learning rate, warming up to $1 \times 10^{-4}$, for a fraction of 0.01 of total training steps, then linearly decay to 0 over 25 epochs. Batch size is 32. Cross entropy loss is used to train the model. Since there are more non-matching poses than matching poses in a pair of given frames, we use Pytorch's WeightedRandomSampler to sample from matching and non-matching classes equally, accounting for class imbalance. When assigning a track ID to a pose, we choose the maximum match score from the previous 4 timesteps. All models are trained on 1 NVIDIA GTX 1080Ti GPU.

\paragraph{Inference} The detector processes images with a batch size of 1. The detections are fed to the pose estimator, which processes all of the bounding box detections for a frame in a single batch. Flip testing is used. Temporal OKS is computed for every frame with an OKS threshold of 0.35. The bounding box scores are ignored when computing OKS. Bounding boxes are thresholded at a minimum confidence score of 0.2, and keypoints are thresholded at a minimum confidence score of 0.1. We found decreasing the bounding box confidence and keypoint thresholds to 0 did not improve AP, but hurt runtime. Boxes are enlarged by factor $\alpha = 1.25$. All code is written in Python, and we use 1 NVIDIA GTX 1080ti. As done by \cite{HRNet, girdhar2018detect}, we train on the PoseTrack 2017 Train and Validation sets before evaluating on the heldout Test Set.

\paragraph{Details of the Tracking Pipeline Analysis} The ablation studies from \ref{sec:pipeline_study} were conducted using the predicted keypoints and predicted boxes with our best model on the PoseTrack 2018 Validation Set. Match accuracy, the metric we use in Table \ref{Tab:embeddings} is similar to $1 - \% IDSW$, i.e. the IDs which are not switched. The methods would be in the same order if measured with IDSW.

\subsection{Architecture Details} \label{sec:details}

\paragraph{Detector and Estimator} We use the implementation of the COCO-pretrained Hybrid Task Cascade Network \cite{chen2019hybrid} with Deformable Convolutions and Multi Scale predictions from \cite{mmdetection}. For our pose estimator, we use the most accurate model from \cite{HRNet}, HRNetW48-384x288.

\paragraph{Transformer Matching Network} We use an effective image resolution of $24 \times 18$ for a total of $432$ unique Position tokens. There are $|\mathcal{K}|=15$ Type tokens and $4$ Segment Tokens.

Each pair of poses has $2|\mathcal{K}|$ tokens total. These are projected to embeddings with dimension $[2|\mathcal{K}|, H]$, where $H=128$ is the transformer hidden size (this is also the transformer intermediate size). The sum of each token's embedding is input to our Transformer Matching Network. The network's backbone consists of 4 transformers in series, each with 4 attention heads. We use a probability of 0.1 for dropout, applying it throughout our Network as \cite{BERT}.  Weights are initialized from a standard normal, $ \mathcal{N}(0, 0.02)$.  The output is pooled, then fed to a binary classification layer, $[H, 2]$. The network has a total of 0.41M parameters, we adapt code from \cite{HuggingFace}. \ref{tab:transformerlayers} gives details of our transformer, which follows the original architecture. The inputs are the hidden states, $[B, 2|\mathcal{K}|, H]$, where $B$ is batch size, and an attention mask, $[B, 1, 1, 2|\mathcal{K}|]$. The extra dimensions in the attention mask are for broadcasting in matrix multiplication. The FLOP counts for our Transformer Matching Network are in Table \ref{tab:flops}.

\begin{table}[t]
    \centering
    \begin{adjustbox}{width=0.5\textwidth}
        \begin{tabular}{c c c c}
            \toprule
            element 1 & op & element 2 & output \\
            \midrule
             hidden states $[32, 30, 128]$ & x & $W^Q ~[128, 128]$ & $Q ~[32, 30, 128]$ \\
             hidden states $[32, 30, 128]$ & x & $W^K ~[128, 128]$ & $K ~[32, 30, 128]$ \\
             hidden states $[32, 30, 128]$ & x & $W^V ~[128, 128]$ & $V ~[32, 30, 128]$ \\
             $Q ~[32, 30, 128]$ & resize &  -   & $Q_{multi} [32, 4, 30, 32]$ \\
             $K ~[32, 30, 128]$ & resize &  -   & $K_{multi} [32, 4, 32, 30]$ \\
             $V ~[32, 30, 128]$ & resize &  -   & $V_{multi} [32, 4, 30, 32]$ \\
             $A_{scores}~[32, 4, 30, 30]$ & + & attention mask & $A_{scores}'~[32, 4, 30, 30]$\\
             $A_{scores}'~[32, 4, 30, 30]$ & softmax & - & $A_{probs}~[32, 4, 30, 30]$ \\
             $A_{probs}~[32, 4, 30, 30]$ & dropout & - & $A_{probs}~[32, 4, 30, 30]$ \\
             $A_{probs}~[32, 4, 30, 30]$ & x & $V_{multi} [32, 4, 30, 32]$ & context $[32, 4, 30, 32]$ \\
             context $[32, 4, 30, 32]$ & resize & - & context $[32, 30, 128]$ \\
             \bottomrule
        \end{tabular}
        \label{tab:transformerlayers}
    \end{adjustbox}
    \caption{A look inside our transformer. 32 is the batch size. x is matrix multiplication., $Q, K, V$ are the query, key, and value, respectively. $W^*$ are the learned weights corresponding to the query, key, or value. ``multi" refers to a multi-headed representation. $A_{scores}$ are the raw attention scores, and $A_{probs}$ is the distribution of attention scores resulting from the softmax operation.}
\end{table}{}

\begin{table}[]
    \centering
    \begin{adjustbox}{width=0.48\textwidth}
    \begin{tabular}{l c c}
         \toprule
         Network Module & Parameters (M) & FLOPS (M) \\
         \midrule
         Embeddings & 0.06 & 0.35 \\
         Transformers (x4) & 0.40 & 5.84 \\
         Pooler & 0.02 & 0.015 \\
         Classifier & 0.01 & 0.02 \\
         Transformer Matching Network & 0.41 & 6.20 \\
         \hdashline
         GCN & 0.1 & 1.30 \\
         Optical Flow & 38.7 & $52.7 \times 10^3$ \\
         \bottomrule
    \end{tabular}
    \end{adjustbox}
    \caption{FLOP and parameter comparison of our Transformer Matching Network to alternative tracking methods. The first four rows give details of each component of our network. (M) indicates millions. Its computational cost is similar to a GCN, only amounting to 1ms increase on the GPU, and much more efficient than Optical Flow. As we showed earlier, our method is more accurate than both alternatives.}
    \label{tab:flops}
\end{table}{}

\begin{figure*}[t!]
    \centering
    \begin{adjustbox}{width=1.0\textwidth}
        \includegraphics{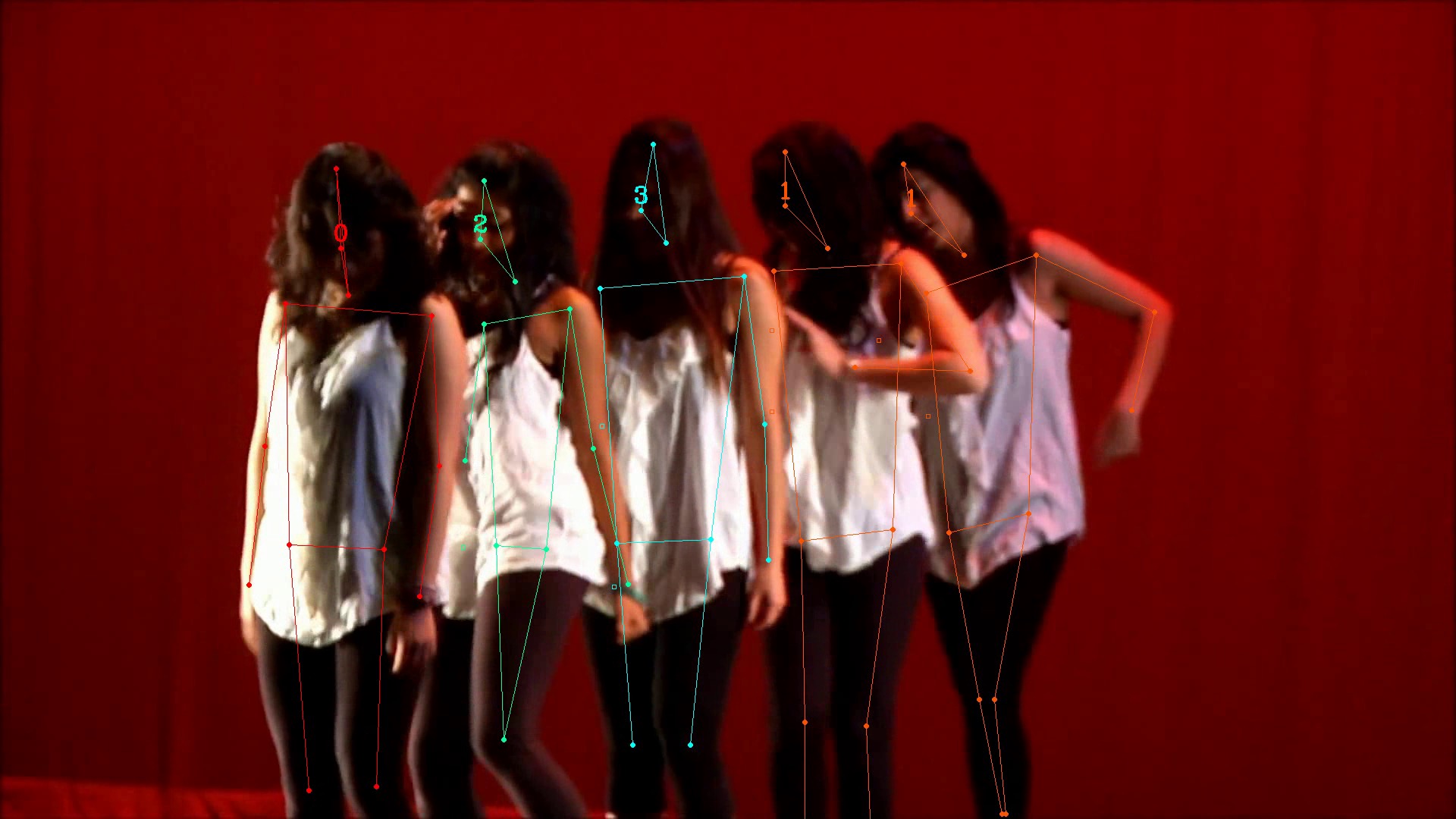}
        \includegraphics{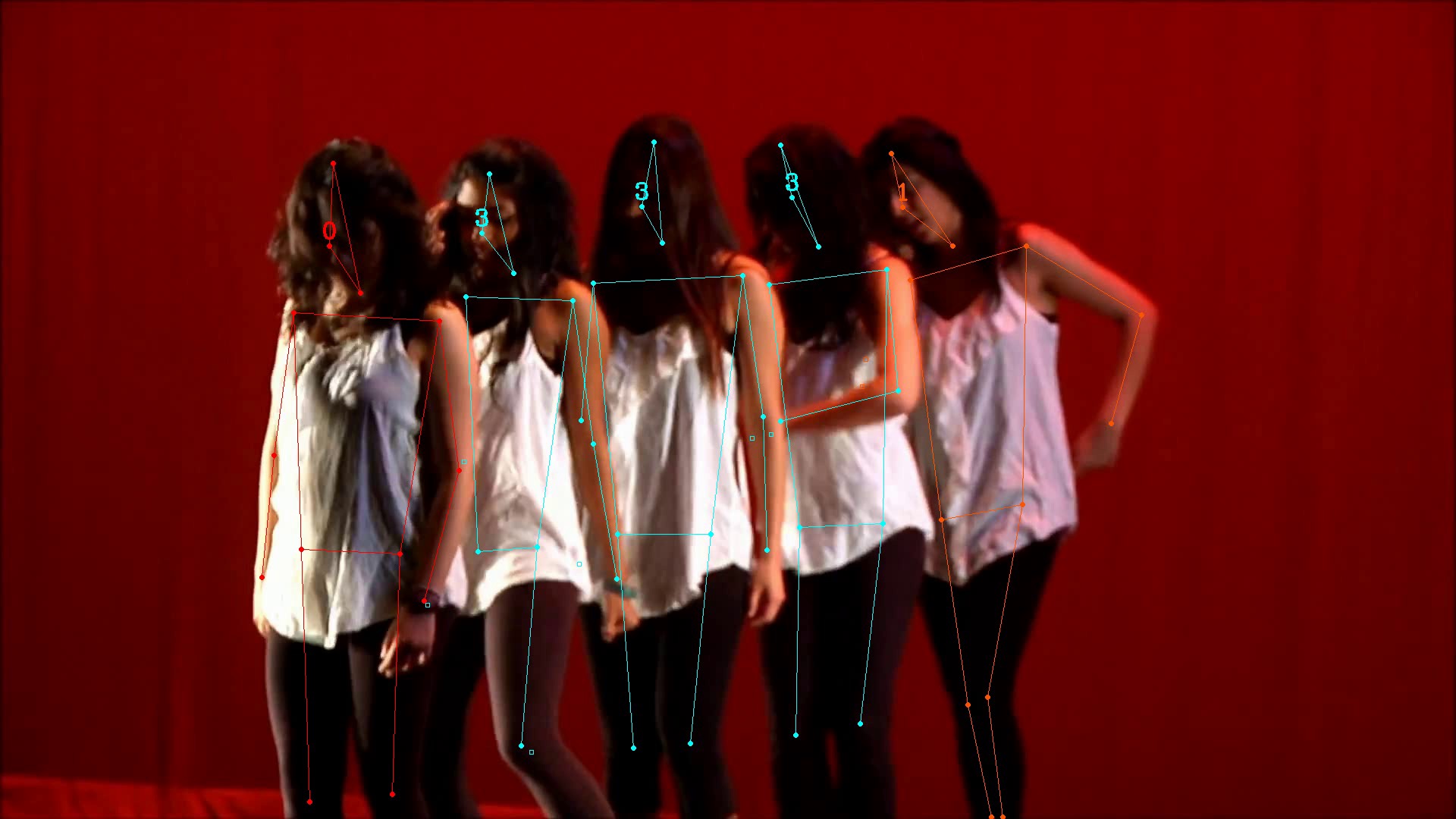}
        \includegraphics{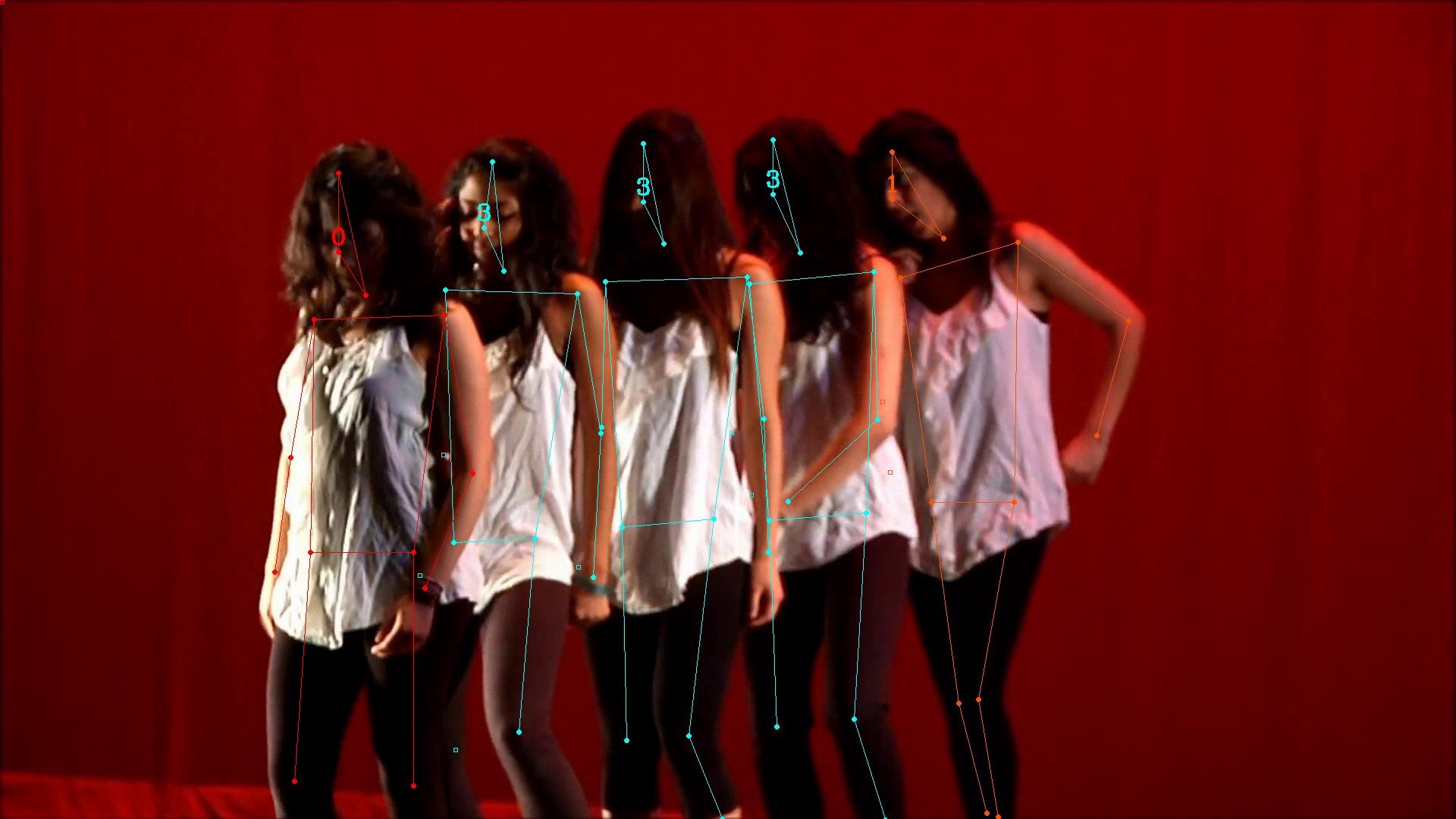}
        \includegraphics{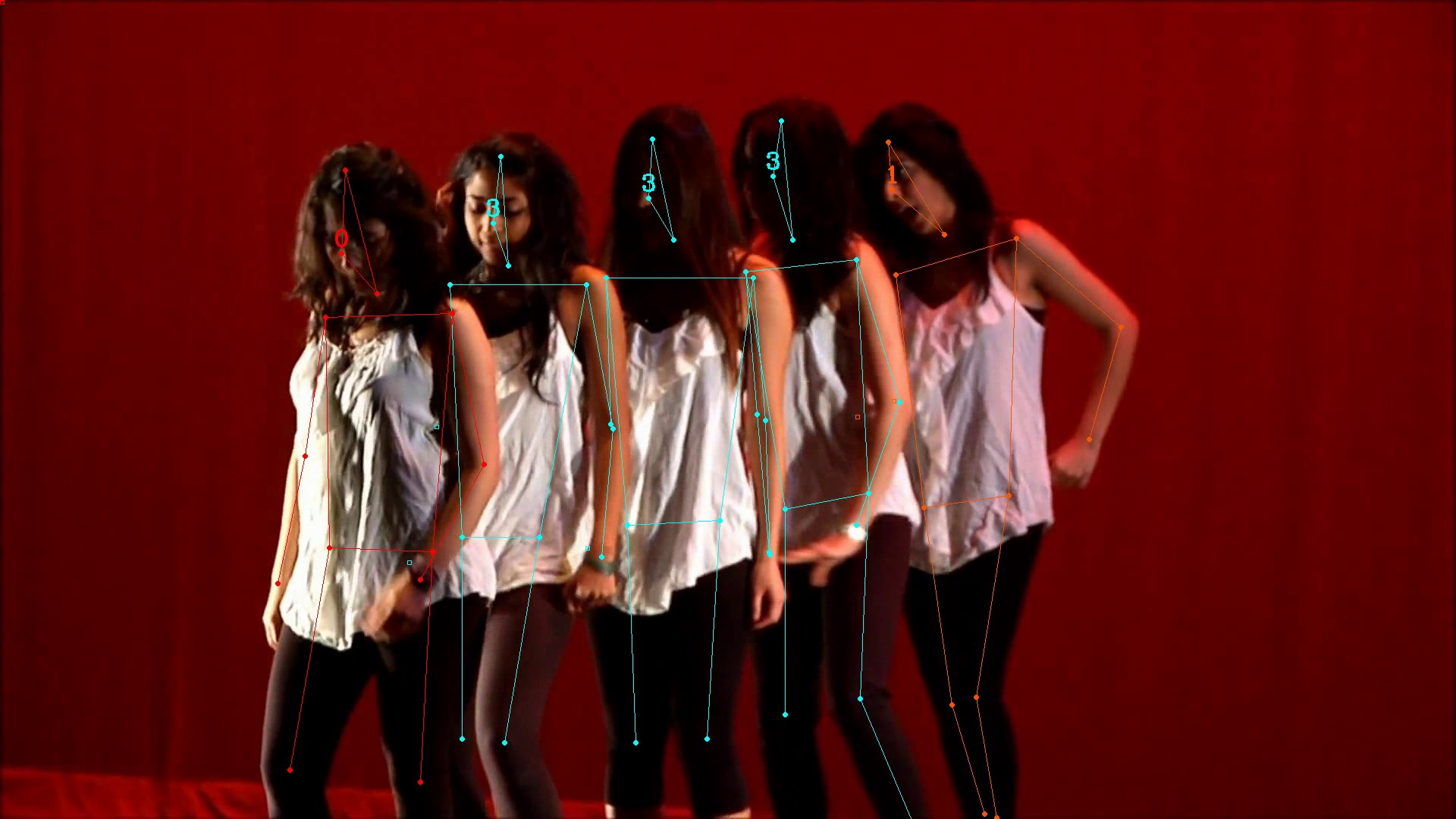}
    \end{adjustbox}
    \tiny
    \begin{tabularx}{\textwidth}{X}
        \phantom{blabla} \\
    \end{tabularx}{}
    \begin{adjustbox}{width=1.0\textwidth}
        \includegraphics{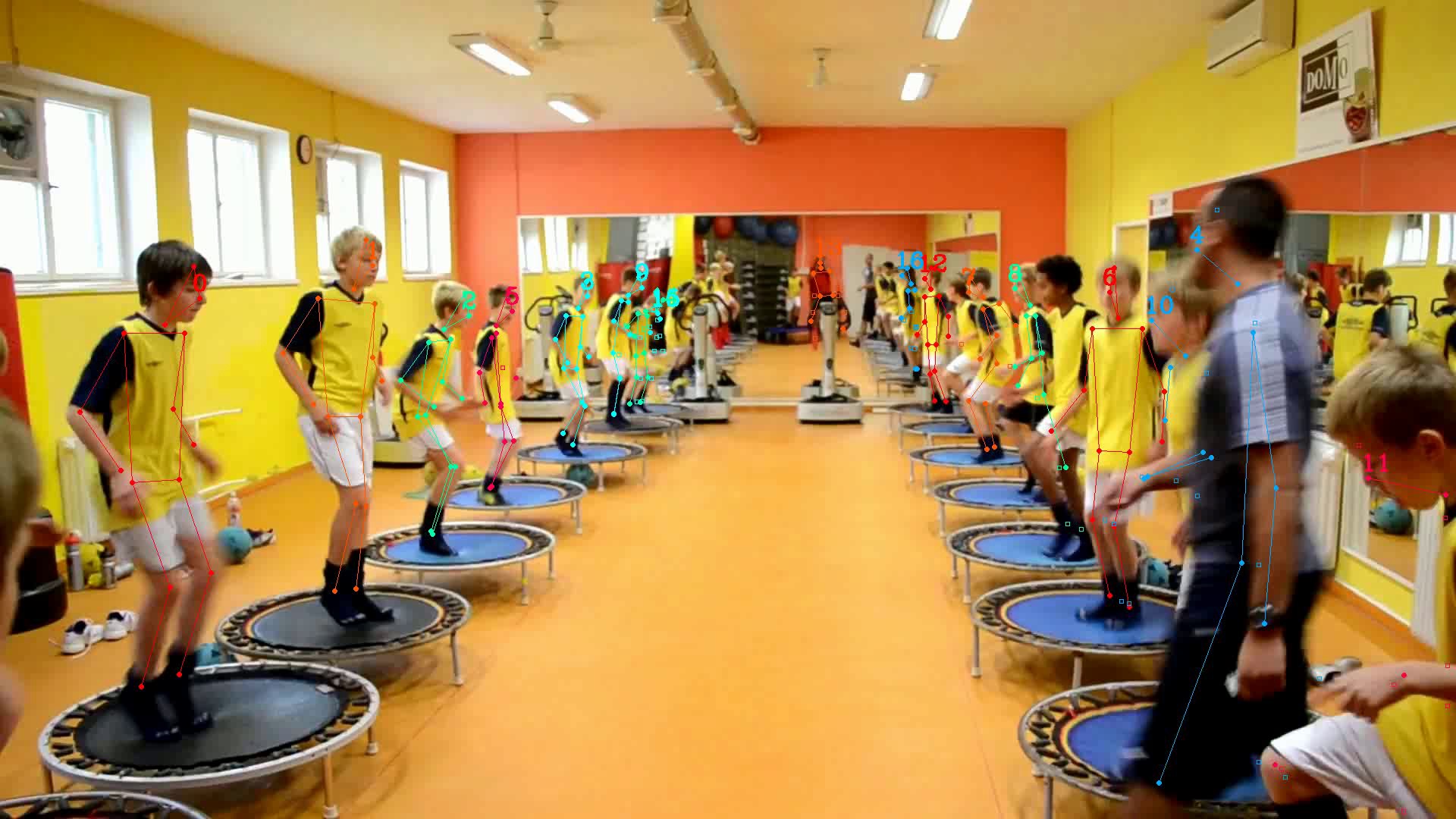}
        \includegraphics{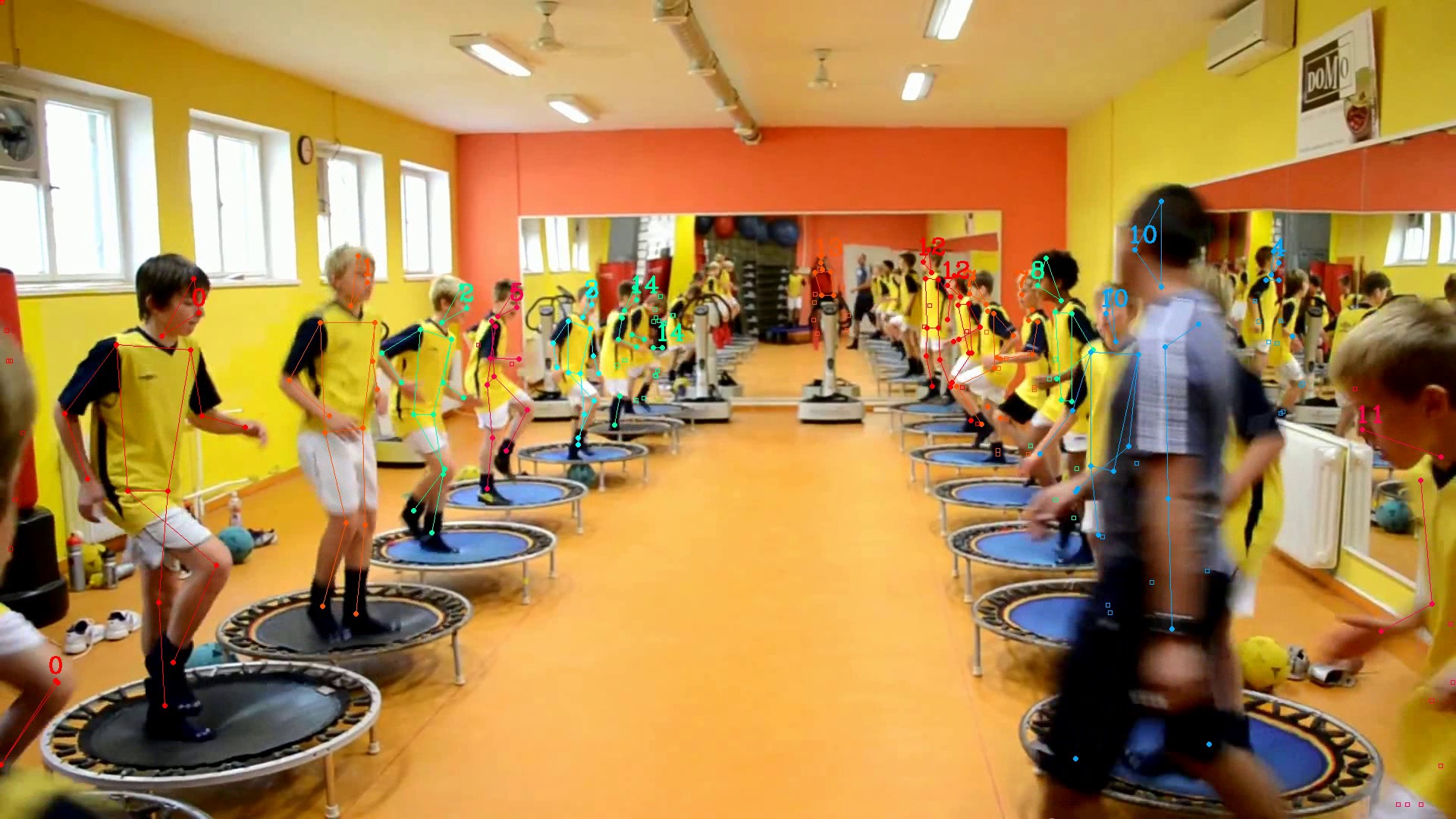}
        \includegraphics{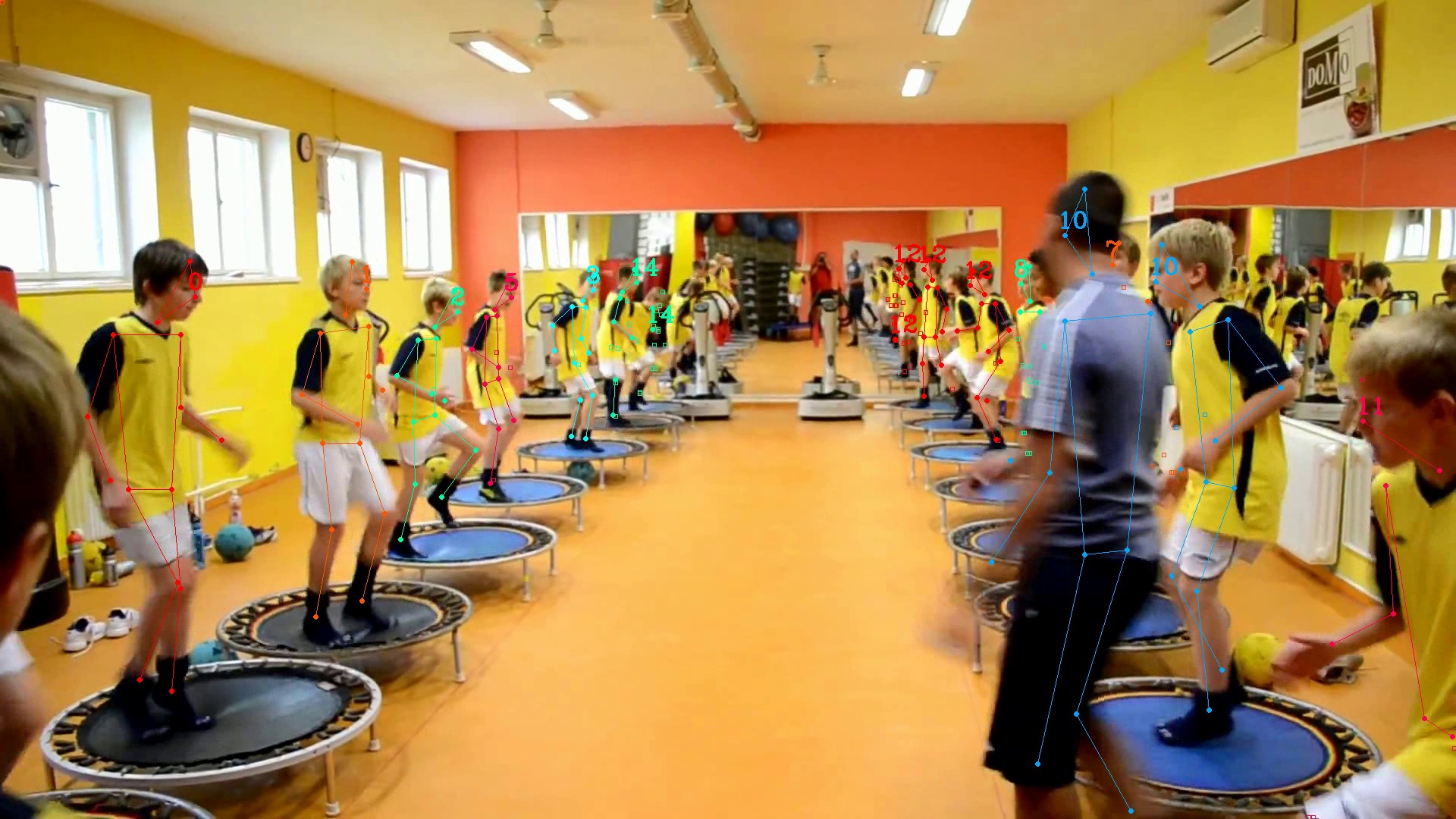}
        \includegraphics{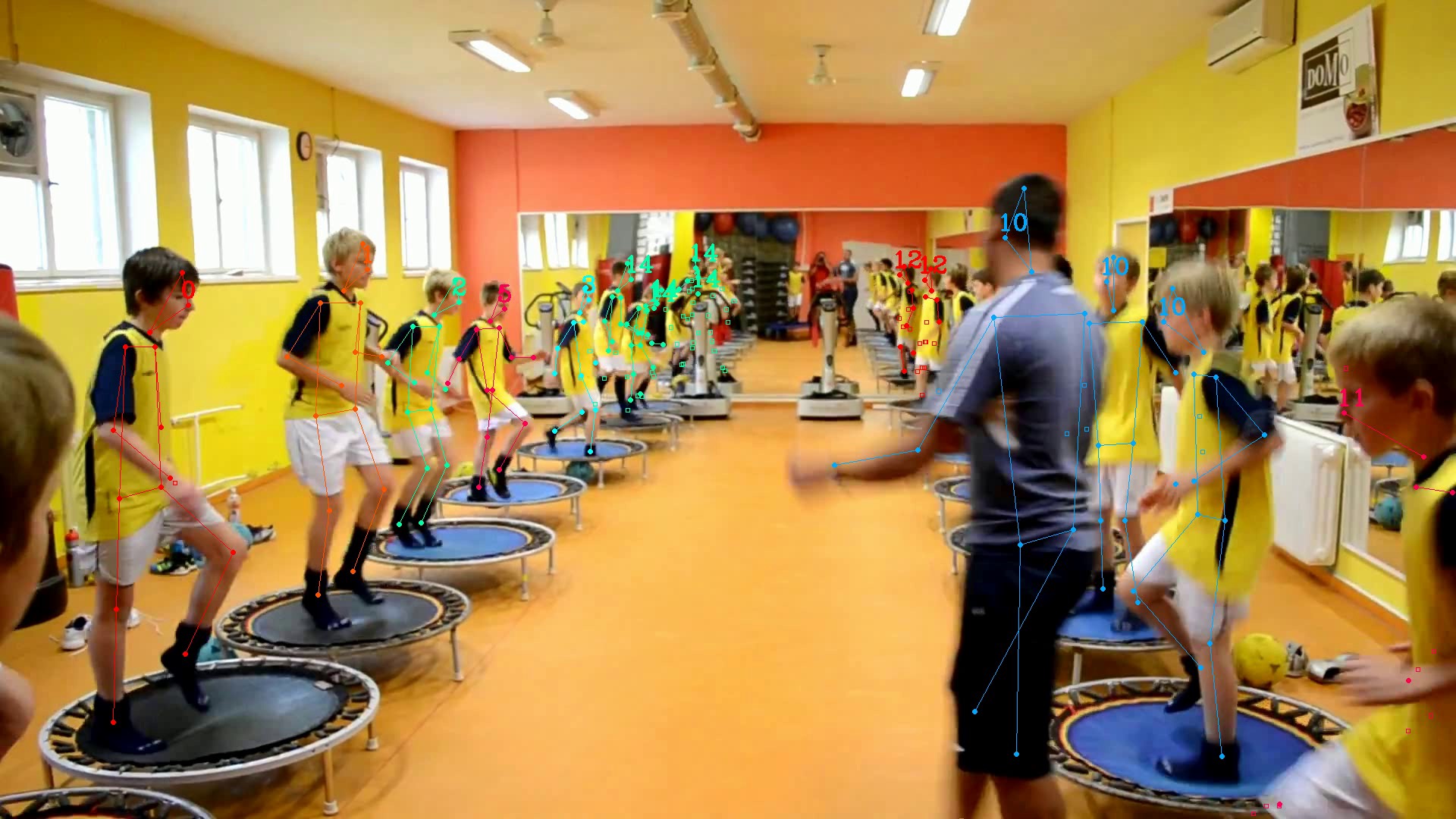}
    \end{adjustbox}
    \caption{Two videos which highlight the limitations of our model. In the top example, the individuals are very near each other and are moving in a synchronized fashion. Thus, our model incorrectly ids people in the middle of the group. In the bottom row, a man walks in front of boys on trampolines. They are occluded for a few frames and are given incorrect ids after he walks away from them. Also, some of the individuals in the back are given incorrect ids because they are small, in close proximity, and moving in similar fashions.}
    \label{fig:limitations}
\end{figure*}{}

\FloatBarrier

We also give details about the other tracking methods we compare to in Table \ref{tab:idsw}. Though our method is slightly more computationally expensive than the GCN, it is much more accurate. Both Transformers and the GCN are far less computationally expensive than Optical Flow.

\paragraph{CNN Pose Entailment Networks} The input is projected to 64 channels in the first layer of the CNN. All convolutions use kernel size 3 and padding 1. BatchNorm is applied after each convolutional layer. The input is downsampled via a maxpooling operation with a stride of 2. The number of filters are doubled after downsampling. Two Linear layers complete the network. The hidden size is dependent on the resolution of the input image. The second layer outputs a binary classification, corresponding to the likelihood of the poses being a match or non-match.

The number of convolutions layers is equal to $log_2(n) - 1$, where $n$ is the long edge of the input image. The batch size, learning rate, and number of training epochs are the same as those we used for the transformers. We experimented with other learning rates, but did not see improvement.

The scheme to color the ``visual tokens" is accomplished by fixing the Hue and Saturation, then adjusting the Value via a linear interpolation from $0-100\%$ in increments of $\frac{100}{|\mathcal{K}|}$.

\subsection{Limitations} Our approach can struggle to track people who are in close proximity and are moving in similar patterns. This is similar to how CNNs struggle with people in close proximity who look visually similar, such as the case where they are wearing the same uniform. Another challenging case for our model is people who are hidden for long periods of time. It is difficult to re-identify them without visual features, and we would need to take longer video clips into context than we currently do to successfully re-identify these individuals. We visualize both these failure modes in Figure \ref{fig:limitations}.

\newpage

{}

\end{document}

%% file: macros.tex
\newcommand{\trackname}{KeyTrack}
\newcommand{\refineshort}{TOKS}
\newcommand{\comment}[1]{}

\definecolor{myorchid}{RGB}{150,10,30}
\definecolor{mygreen}{RGB}{10,120,10}

%% file: sections/abstract.tex
\begin{abstract}
  Pose tracking is an important problem that requires identifying unique human pose-instances and matching them temporally across
  different frames of a video. However, existing pose tracking methods are unable to accurately model temporal relationships and require significant computation, often computing the tracks offline. We present an efficient multi-person pose tracking method, \trackname, that only relies on keypoint information without using any RGB or optical flow information to track human keypoints in real-time. Keypoints are tracked using our Pose Entailment method, in which, first, a pair of pose estimates is sampled from different frames in a video and tokenized. Then, a Transformer-based network makes a binary classification as to whether one pose temporally follows another. Furthermore, we improve our top-down pose estimation method with a novel, parameter-free, keypoint refinement technique that improves the keypoint estimates used during the Pose Entailment step. We achieve state-of-the-art results on the PoseTrack'17 and the PoseTrack'18 benchmarks while using only a fraction of the computation required by most other methods for computing the tracking information.

\end{abstract}

%% file: sections/intro.tex
\section{Introduction}
Multi-person Pose Tracking is an important problem for human action recognition and video understanding. It occurs in two steps: first, estimation, where keypoints of individual persons are localized; second, the tracking step, where each keypoint is assigned to a unique person. Pose tracking methods rely on deep convolutional neural networks for the first step~\cite{toshev2014deeppose, tompson2015efficient, yang2017learning, wei2016convolutional}, but approaches in the second step vary. This is a challenging problem because tracks must be created for each unique person, while overcoming occlusion and complex motion. Moreover, individuals may appear visually similar because they are wearing the same uniform. It is also important for tracking to be performed online. Commonly used methods, such as optical flow and graph convolutional networks (GCNs) are effective at modeling spatio-temporal keypoint relationships~\cite{HRNet},~\cite{ning2019lighttrack}, but are dependent on high spatial resolution, making them computationally costly. Non-learning based methods, such as spatial consistency, are faster than the convolution-based methods, but are not as accurate.

\begin{figure}[t]
    \centering
    \includegraphics[width=0.47\textwidth]{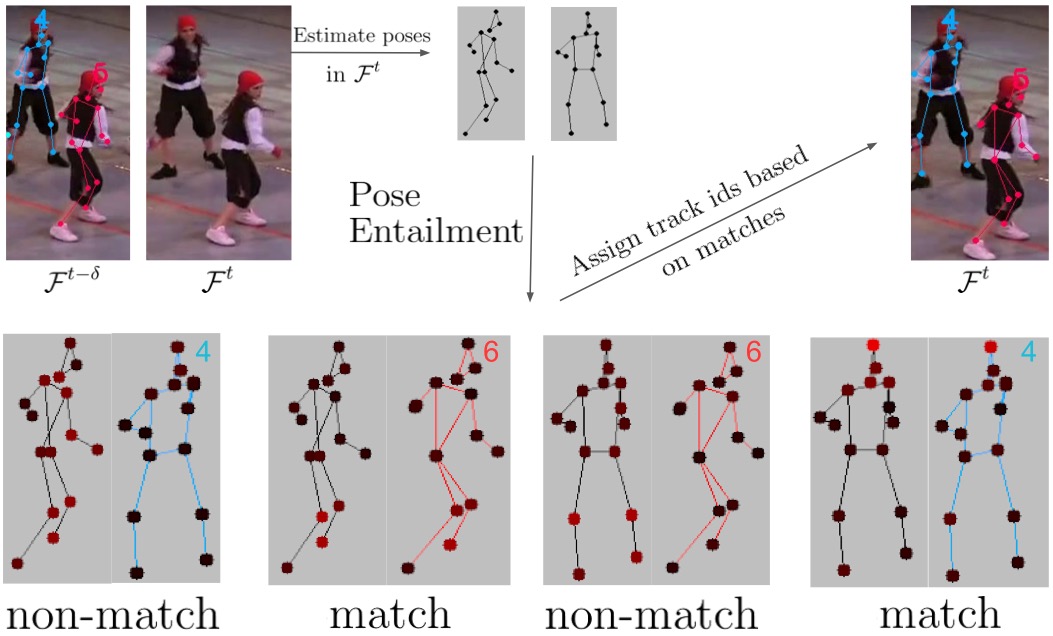}
    \caption{\textbf{They look alike, how do we decide who\textquotesingle s who?} In the Pose Entailment framework, given a video frame, we track individuals by comparing pairs of poses, using temporal motion cues to determine who's who.  Using a novel tokenization scheme to create pose pair inputs interpretable by Transformers \cite{vaswani2017attention}, our network divides its attention equally between both poses in {\em matching pairs}, and focuses more on a single pose in {\em non-matching} pairs because motion cues between keypoints are not present. We visualize this above; bright red keypoints correspond to high attention.}
    \vspace{-0.2in}
\end{figure}

To address the above limitations, we propose an efficient pose tracking method, \trackname, that leverages temporal relationships to improve multi-person pose estimation and tracking. Hence, \trackname\ follows the {\em tracking by detection} approach by first localizing humans, estimating human pose keypoints and then encoding the keypoint information in a novel {\em entailment} setting using transformer building blocks~\cite{vaswani2017attention}. Similar to the textual entailment task where one has to predict if one sentence follows one another, we propose the {\em Pose Entailment} task, where the model learns to make a binary classification if two keypoint poses temporally follow or entail each other. Hence, rather than extracting information from a high-dimensional image representation using deep CNNs, we extract information from a sentence of 15 tokens, and each token corresponds to a keypoint on a pose. Similar to how BERT tokenizes words~\cite{devlin2018bert}, we propose an embedding scheme for pose data that captures spatio-temporal relationships and feed our transformer network these embeddings. Since these embeddings contain information beyond spatial location, our network outperforms convolution based approaches in terms of accuracy and speed, particularly at very low resolutions. 
 
Additionally, in order to improve the keypoint estimates used by the transformer network, we propose a Temporal Object Keypoint Similarity (\refineshort) method. \refineshort\ refines the pose estimation output by augmenting missed detections and thresholding low quality keypoint estimates using a keypoint similarity metric. \refineshort\ adds no learned parameters to the estimation step, and is superior to existing bounding box propagation methods that often rely on NMS and optical flow.  \trackname\ makes the following contributions: 
 
 \textbf{1.} \trackname\ introduces Pose Entailment, where a binary classification is made as to whether two poses from different timesteps are the same person. We model this task in a transformer-based network which learns temporal pose relationships even in datasets with complex motion. Furthermore, we present a tokenization scheme for pose information that allows transformers to outperform convolutions at low spatial resolutions when tracking keypoints.
 
 \textbf{2.} \trackname\ introduces a temporal method for improving keypoint estimates. \refineshort ~is more accurate than bounding box propagation, faster than a detector ensemble, and does not require learned parameters. 

Using the above methods, we develop an efficient multi-person pose tracking pipeline which sets a new SOTA on the PoseTrack test set. We achieve 61.2\% tracking accuracy on the PoseTrack'17 Test Set and 66.6\% on the PoseTrack'18 Val set using a model that consists of just 0.43M parameters in the tracking step, making this portion of our pipeline 500X more efficient than than the leading optical flow method~\cite{HRNet}. Our training is performed on a single NVIDIA 1080Ti GPU. Not reliant on RGB or optical flow information in the tracking step, our model is suitable to perform pose tracking using other non-visual pose estimation sensors that only provide 15 keypoints for each person~\cite{alarifi2016ultra}.

%% file: sections/related.tex
\section{Related Work}
\label{sec:related}

We are inspired by related work on pose estimation and tracking methods, and recent work on applying the transformer network to video understanding.

\begin{figure*}[t]
    \centering
    \includegraphics[width=\textwidth]{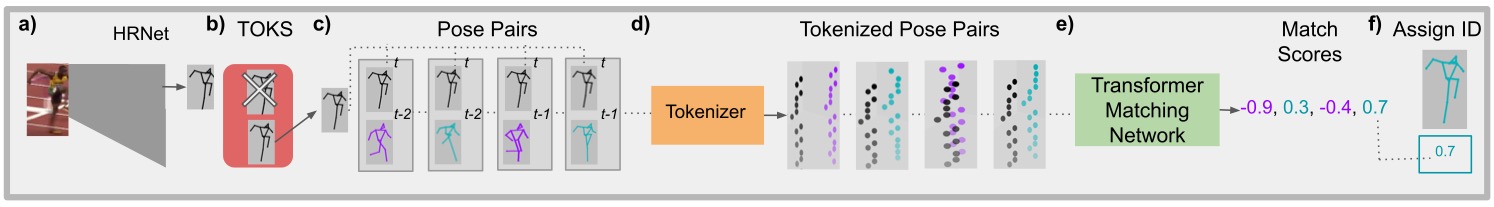}
    \caption{\label{fig:overview} \textbf{a)} Keypoints are estimated with HRNet. \textbf{b)} TOKS improves detection accuracy. \textbf{c)} Pose pairs are collected from multiple past timesteps. Poses of the same color have the same track id, the color black indicates the track id is unknown. \textbf{d)} Each pair is tokenized independently from the other pairs. \textbf{e)} Our Transformer Matching Network calculates match scores independently for each pair. \textbf{f)} The maximum match score is greedily chosen and the corresponding track id is assigned.}
\end{figure*}
\paragraph{Pose estimation} Early work on pose estimation has focused on using graphical models to learn spatial correlations and interactions between various joints~\cite{andriluka2009pictorial, felzenszwalb2005pictorial}. 
These models often perform poorly due to occlusions and long range temporal relationships, which need to be explicitly modeled in this framework~\cite{dantone2013human, sigal2006measure, wang2008multiple}. 
More recent work involves using convolutional neural networks (CNNs) to directly regress cartesian co-ordinates of the joints~\cite{toshev2014deeppose} or to generate heatmaps of the probability of a joint being at a specific location~\cite{tompson2015efficient, yang2017learning, wei2016convolutional}.
A majority of the convolutional approaches can be classified into top-down and bottom-up methods -- the top-down methods use a separate detection step to identify person candidates~\cite{he2017mask, papandreou2017towards, chen2018cascaded, huang2017coarse, papandreou2017towards}. The single person pose estimation step is then performed on these person candidates. Bottom-up methods calculate keypoints from all candidates and then correlate these keypoints into individual human joints~\cite{xia2017joint, hwang2019pose}. The latter method is more efficient since all keypoints are calculated in a single step; however, the former is more accurate since the object detection step limits the regression boundaries. However, top-down methods work poorly on small objects and recent work (HRNet)~\cite{HRNet} uses parallel networks at different resolutions to maximize spatial information. PoseWarper~\cite{bertasius2019learning} uses a pair of labeled and unlabeled frames to predict human pose by learning the pose-warping using deformable convolutions. Finally, since the earliest applications of deep learning to pose estimation~\cite{toshev2014deeppose}, iterative predictions have improved accuracy. Pose estimation has shown to benefit from cascaded predictions \cite{chen2018cascaded} and pose-refinement methods~\cite{fieraru2018learning, moon2019posefix} refine the pose estimation results of previous stages using a separate post-processing network. In that spirit, our work, \trackname\ relies on HRNet to generate keypoints and refines keypoint estimates by temporally aggregating and suppressing low confidence keypoints with \refineshort\ instead of commonly used bounding box propagation approaches.

\begin{center}{}
\begin{table}
\renewcommand{\arraystretch}{1.2}
\begin{adjustbox}{max width=0.48\textwidth}
    \begin{tabular}{llll|c}
        \toprule
         \multicolumn{1}{c}{Method} & 
         \multicolumn{1}{c}{Estimation} & 
         \multicolumn{1}{c}{Detection Improvement} & 
         \multicolumn{1}{c}{Tracking} & \\
         \midrule
         Ours & HRNet & \bf Temporal OKS & \bf Pose Entailment & \\
         HRNet~\cite{HRNet} & HRNet & BBox Prop. & Optical Flow &
         \parbox[t]{2mm}{\multirow{4}{*}{\rotatebox[origin=c]{90}{Top-Down}}} \\
         POINet~\cite{Ruan:2019:PPO:3343031.3350984} & VGG, T-VGG & - & Ovonic Insight Net \\
         MDPN \cite{Guo_2019} & MDPN & Ensemble & Optical Flow & \\
         LightTrack~\cite{ning2019lighttrack} & Simple Baselines & Ensemble/BBox Prop. & GCN & \\
         ProTracker~\cite{girdhar2018detect} & 3D Mask RCNN & - & IoU & \\
         \midrule
         Affinity Fields~\cite{raaj2019efficient} & VGG/STFields & - & STFields & \parbox[t]{2mm}{\multirow{3}{*}{\rotatebox[origin=c]{90}{Bottom-Up}}}  \\
         STEmbeddings~\cite{jin2019multi} & STEmbeddings & - & STEmbeddings & \\
         JointFlow & Siamese CNN & - & Flow Fields & \\
         
        \bottomrule
    \end{tabular}
\end{adjustbox}
\caption{How different approaches address each step of the Pose Tracking problem. Our contributions are in bold.}
\label{table:relatedwork}
\end{table}
\end{center}

\vspace{-1cm}

\paragraph{Pose tracking Methods} Pose tracking methods assign unique IDs to individual keypoints, estimated with techniques described in the previous subsection, to track them through time~\cite{PoseTrack, insafutdinov2017arttrack, iqbal2017posetrack, PoseTrack2017Leaderboard}. Some methods perform tracking by learning spatio-temporal pose relationships across video frames using convolutions ~\cite{wang2019learning, Ruan:2019:PPO:3343031.3350984, ning2019lighttrack}. \cite{Ruan:2019:PPO:3343031.3350984}, in an end-to-end fashion, predicts track ids with embedded visual features from its estimation step, making predictions in multiple temporal directions. \cite{ning2019lighttrack} uses a GCN to track poses based on spatio-temporal keypoint relationships. These networks require high spatial resolutions. In contrast, we create keypoint embeddings from the keypoint's spatial location and other information. This makes our network less reliant on spatial resolution, and thus more efficient, and gives our network the ability to model more fine-grained spatio-temporal relationships. 

Among non-learned tracking methods, optical flow is effective. Here, poses are propagated from one frame to the next with optical flow to determine which pose they are most similar to in the next frame \cite{HRNet, Guo_2019}. This improves over spatial consistency, which measures the IoU between bounding boxes of poses from temporally adjacent frames ~\cite{girdhar2018detect}. Other methods use graph-partitioning based approaches to group pose tracks ~\cite{insafutdinov2017arttrack, iqbal2017posetrack, jin2017towards}. Another method, PoseFlow~\cite{xiu2018pose}, uses inter/intra-frame pose distance and NMS to construct pose flows. However, our method does not require hard-coded parameters during inference, this limits the ability of non-learned methods to model scenes with complex motion and requires time-intensive manual tuning. Table~\ref{table:relatedwork} shows top-down methods similar to our work as well as competitive bottom-up methods.

\paragraph{Transformer Models}
Recently, there have been successful implementations of transformer based models for image and video
input modalities often substituting convolutions and recurrence mechanisms. These methods can efficiently
model higher-order relationships between various scene elements unlike pair-wise methods~\cite{dai2017detecting, hu2016modeling, santoro2017simple, xu2019spatial}.  
They have been applied for image classification~\cite{ramachandran2019stand}, visual question-answering~\cite{li2019visualbert, lu2019vilbert, tan2019lxmert, zhou2019unified}, action-recognition~\cite{huangdynamic, ma2018attend}, video captioning~\cite{sun2019contrastive, zhou2018end} and other video problems.  Video-Action Transformer~\cite{girdhar2019video} solves the action localization problem using transformers by learning the context and interactions for every person in the video.
BERT~\cite{BERT} uses transformers by pretraining a transformer-based network in a multi-task transfer learning scheme over the unsupervised tasks of predicting missing words or next sentences. Instead, in a supervised setting, \trackname\ uses transformers to learn spatio-temporal keypoint relationships for the visual problem of pose tracking.

%% file: sections/methods.tex
\section{Method}

\begin{figure*}[t]
    \centering
    \includegraphics[width=0.95\textwidth]{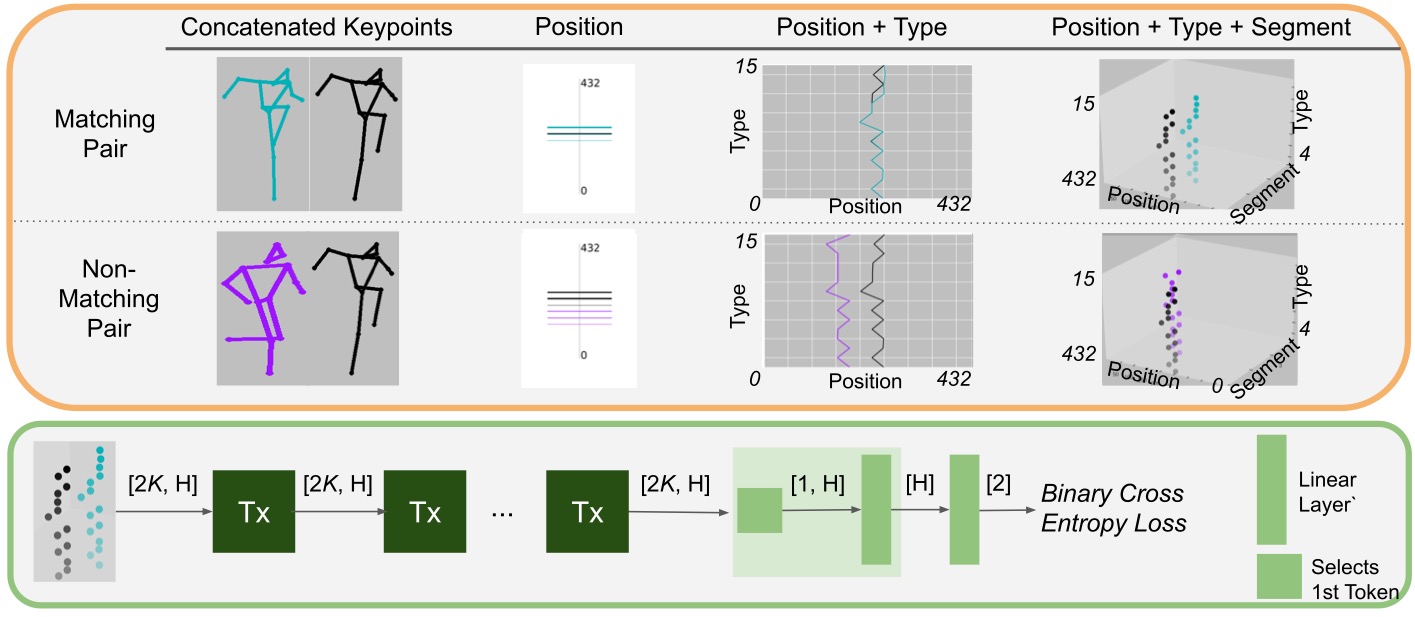}
    \caption{\label{fig:tokens} \textbf{Orange box:}  Visualizations to intuitively explain our tokenization. In the Position column, the matching poses are spatially closer together than the non-matching ones. This is because their spatial locations in the image are similar. The axis limit is $432$ because the image has been downsampled to $width * height = 432$. In the following column, the matching contours are similar, since the poses are in similar orientations. The Segment axis in the last column represents the temporal distance of the pair. \textbf{Green box:} A series of transformers (Tx) compute self-attention, extracting the temporal relationship between the pair. Binary classification follows.}
\end{figure*}

\subsection{Overview of Our Approach}
We now describe the keypoint estimation and tracking approach used in \trackname\ as shown in Figure~\ref{fig:overview}.
For frame $\mathcal{F}^t$ at timestep $t$, we wish to assign a track id to the $i$th pose $p^{t,i} \in \mathcal{P}^t$. First, each of the pose's $k^j \in \mathcal{K}$ keypoints are detected. This is done by localizing a bounding box around each pose with an object detector and then estimating keypoint locations in the box. Keypoint predictions are improved with temporal OKS (\refineshort). Please see \ref{estimation} for more details. From here, this pose with no tracking id, $p^{t,i}_{\diameter}$, is assigned its appropriate one. This is based on the pose's similarity to a pose in a previous timestep, which has an id, $p^{t-\delta,j}_{id}$. Similarity is measured with the match score, $m^{t-\delta,j}_{id}$, using Pose Entailment (\ref{pose_entailment}).

False negatives are an inevitable problem in keypoint detection, and hurt the downstream tracking step because poses with the correct track id may appear to be no longer in the video. We mitigate this by calculating match scores for poses in not just one previous frame, but multiple frames $ \{ \mathcal{F}^1, \mathcal{F}^2, ~ ... ~ \mathcal{F}^\delta \}$. Thus, we compare to each pose $p^{t-d,j}_{id}$ where $1 \leq d \leq \delta$ and $1 \leq j \leq |\mathcal{P}^{t-d}|$. In practice, we limit the number of poses we compare to in a given frame to the $n$ spatially nearest poses. This is just as accurate as comparing to everyone in the frame and bounds our runtime to $O(\delta n)$. This gives us a set of match scores $\mathcal{M}$, and we assign $p^{t,i}_{\diameter}$ the track id corresponding to the maximum match score $m^* = max(\mathcal{M})$, where $id^* = m^*_{id}$. Thus, we assign the tracking id to the pose, $p^{t,i}_{id^*}$.

\subsection{Pose Entailment} \label{pose_entailment}

To effectively solve the multi-person pose tracking problem, we need to understand how human poses move through time based on spatial joint configurations as well as in the presence of multiple persons and occluding objects. Hence, to correctly learn temporal transformations through time, we need to learn if a pose in timestep $t$, can be inferred from timestep $t-1$.  Textual entailment provides us with a similar framework in the NLP domain where one needs to understand if one sentence can be implied from the next. More specifically, the textual entailment model classifies whether a \emph{premise} sentence implies a \emph{hypothesis} sentence in a sentence pair~\cite{bowman2015large}. The typical approach to this problem consists of first projecting the pair of sentences to an embedding space and then feeding them through a neural network which outputs a binary classification for the sentence pair. 

 Hence, we propose the Pose Entailment problem. More formally, we seek to classify whether a pose in a timestep $p^{t-\delta}$, i.e. the \emph{premise}, and a pose in timestep $p^{t}$, i.e. the \emph{hypothesis}, are the same person. To solve this problem, instead of using visual feature based similarity that incurs large computational cost, we use the set of human  keypoints, $\mathcal{K}$, detected by our pose estimator. It is computationally efficient to use these as there are a limited number of them (in our case $|\mathcal{K}| = 15$), and they are not affected by unexpected visual variations such as lighting changes in the tracking step. In addition, as we show in the next section, keypoints are amenable to tokenization. Thus, during the tracking stage, we use only the keypoints estimated by the detector as our pose representation.

\paragraph{Tokenizing Pose Pairs} 
The goal of tokenization is to transform pose information into a representation that 
facilitates learning spatio-temporal human pose relationships. To achieve this goal, for each pose token, we
need to provide (i) the spatial location of each keypoint in the scene to allow the network to spatially correlate keypoints across frames, (ii) type information of each keypoint (i.e. head, shoulder etc.) to learn spatial joint relationships in each human pose, and finally (iii) the temporal location index for each keypoint within a temporal window $\delta$, to learn temporal keypoint transitions. Hence, we use three different types of tokens for each keypoint as shown in Figure~\ref{fig:tokens}. There are $2$ poses, and thus $2|\mathcal{K}|$ tokens of each type. Each token is linearly projected to an embedding, $E \in \mathbb{R}^{2\mathcal{|K|},H} $ where $H$ is the transformer hidden size. Embeddings are a learned lookup table. We now describe the individual tokens in detail:

\textbf{Position Token:} The absolute spatial location of each keypoint is the \emph{Position} token, $\rho$, and its values fall in the range $[1, w^{\mathcal{F}}h^{\mathcal{F}}]$. In practice, the absolute spatial location of a downsampled version of the original frame is used. This not only improves the efficiency of our method, but also makes it more accurate, as is discussed in \ref{sec:self-attention}. 
A general expression for the Position tokens of poses $p^t$ and $p^{t-\delta}$ is below, where $\rho^{p^t}_j$ corresponds to the Position token of the $jth$ keypoint of $p^t$:

\begin{equation}
       \resizebox{0.4\textwidth}{!}{
        $\{
        \rho^{p^t}_1, \rho^{p^t}_2, ~~... ~~ \rho^{p^t}_\mathcal{|K|},
        ~\rho^{p^{t-\delta}}_1,~ \rho^{p^{t-\delta}}_2, ~~... ~~ \rho^{p^{t-\delta}}_\mathcal{|K|}
        \} $
    } 
\end{equation}

\textbf{Type Token:} The \emph{Type} token corresponds to the unique type of the keypoint: e.g. the head, left shoulder, right ankle, etc... The \emph{Type} keypoints fall in the range $[1, |\mathcal{K}|]$. These add information about the orientation of the pose and are crucial for achieving high accuracy at low resolution, when keypoints have similar spatial locations. 
A general expression for the \emph{Type} tokens of poses $p^t$ and $p^{t-\delta}$ is below, where $j^{p^t}$ corresponds to the \emph{Type} token of the $jth$ keypoint of $p^t$:
\begin{equation}
       \resizebox{0.4\textwidth}{!}{
        $\{
        1^{p^t}, 2^{p^t}, ~~... ~~ |\mathcal{K}|^{p^t},
        ~1^{p^{t-\delta}},~ 2^{p^{t-\delta}}, ~~... ~~ |\mathcal{K}|^{p^{t-\delta}}
        \} $
    } 
\end{equation}

\textbf{Segment Token:} The \emph{Segment} token indicates the number of timesteps the pose is from the current one. The segment token is in range $[1, \delta]$, where $\delta$ is a chosen constant. (For our purposes, we set $\delta$ to be 4.) This also allows our method to adapt to irregular frame rates. Or, if a person is not detected in a frame, we can look back two timesteps, conditioning our model on temporal token value of $2$ instead of $1$. 
\begin{equation}
       \resizebox{0.4\textwidth}{!}{
        $\{
        1^{p^t}, 1^{p^t}, ~~... ~~ 1^{p^t},
        ~\delta^{p^{t-\delta}},~ \delta^{p^{t-\delta}}, ~~... ~~ \delta^{p^{t-\delta}}
        \} $
    } 
\end{equation}

After each token is embedded, we sum the embeddings, $E_{sum} = E_{Position} + E_{Type} + E_{Segment}$, to combine the information from each class of token. This is fed to our Transformer Matching Network.

\begin{figure*}[t!]
\centering
\renewcommand{\arraystretch}{0.95}
\begin{adjustbox}{width=1.0\textwidth}
    \begin{tabular}{l|cc|cccccccc|c}
        \toprule
        \multirow{2}{*}{Tracking Method} & \multirow{2}{*}{Detection Method} &
        AP $\uparrow$ &
        \multicolumn{8}{c}{\% IDSW $\downarrow$ } &
        MOTA $\uparrow$ \\
          & & Total &
         Head & 
         Shou &
         Elb  &
         Wri  &
         Hip  &
         Knee &
         Ankl &
         Total & 
         Total \\         
         \midrule
         \textbf{Pose Entailment} \quad  \quad& \multirow{3}{*}{GT Boxes, GT Keypoints} & \multirow{3}{*}{100} & \bf 0.7 & \bf 0.7 & \bf 0.6 & \bf 0.6 & \bf 0.6 & \bf 0.7 & \bf 0.7 & \bf 0.7 & \bf{99.3} \\
         GCN & & & 1.4 & 1.4 & 1.4 & 1.5 & 1.4 & 1.6 & 1.6 & 1.5 & 98.5 \\
         Optical Flow & & &  1.1 &  1.2 &  1.2 &  1.2 &  1.2 &  1.3 &  1.4 &  1.2 & 98.7 \\
         \hdashline % 
         \textbf{Pose Entailment} & \multirow{3}{*}{GT Boxes, Predicted Keypoints} & \multirow{3}{*}{86.7} & \bf 0.9 &  \bf 0.9 & \bf 0.8 &  \bf 0.8 & \bf 0.7 & \bf 0.8 & \bf 0.8 & \bf 0.8 & \bf 72.2 \\
         GCN & & & 1.6 &  1.6 &  1.6 &  1.6 &  1.3 &  1.5 &  1.4 &  1.5 & 71.6 \\
         Optical Flow & & &  1.2 &  1.2 &  1.2 &  1.1 &  1.0 &  1.1 &  1.1 &  1.1 & 71.8 \\
         \hdashline
         \textbf{Pose Entailment} & \multirow{3}{*}{Predicted Boxes, Predicted Keypoints} & \multirow{3}{*}{81.6} &  \bf 0.9 &  \bf 1.0 &  \bf 0.9 &  \bf 0.8 &  \bf 0.7 &  \bf 0.8 &  \bf 0.8 &  \bf 0.8 & \bf 66.6 \\
         GCN & & &  1.7 &  1.7 &  1.7 &  1.7 &  1.4 &  1.5 &  1.4 &  1.6 & 65.9 \\
         Optical Flow & & & 1.3 &  1.2 &  1.2 &  1.2 &  1.1 &  1.1 &  1.1 &  1.1 & 66.3 \\
         \bottomrule
    \end{tabular}
\end{adjustbox}{}
\caption{Compares accuracy of tracking methods on the PoseTrack 18 Val set, given the same keypoints. GT stands for Ground Truth, ``predicted" means a neural net is used. Lower \% IDSW is better, higher MOTA is better. ``Total" averages all joint scores.}
\label{tab:idsw}
\end{figure*}

\paragraph{Transformer Matching Network:}

The goal of our network is to learn motion cues indicative of whether a pose pair matches. The self-attention mechanism of transformers allows us to accomplish this by learning which temporal relationships between the keypoints are representative of a match. Transformers compute scaled dot-product attention over a set of Queries ($Q$), Keys ($K$), and Values($V$) each of which is a linear projection of the input $E_{sum} \in \mathbb{R}^{2\mathcal{|K|},H}$. We compute the softmax attention with respect to every keypoint embedding in the pair, with the input to the softmax operation being of dimensions $[2|K|, 2|K|]$. In fact, we can generate heatmaps from the attention distribution over the pair's keypoints, as displayed in \ref{sec:heatmap}. In practice, we use multi-headed attention, which leads to the heads specializing, also visualized.

Additionally, we use an attention mask to account for keypoints which are not visible due to occlusion. This attention mask is implemented exactly as the attention mask in \cite{vaswani2017attention}, resulting in no attention being paid to the keypoints which are not visible due to occlusion. The attention equation is as follows, and we detail each operation in a single transformer in Table ~\ref{tab:transformerlayers}:
\begin{equation}
       \resizebox{0.4\textwidth}{!}{
        $
        \text{Attention}(Q, K, V) = \text{softmax}(\frac{QK^T}{\sqrt{d_k}})V
        $
    } 
\end{equation}

After computing self-attention through a series of stacked transformers, similar to BERT, we feed this representation to a Pooler, which ``pools" the input, by selecting the first token in the sequence and then inputting that token into a learned linear projection. This is fed to another linear layer, functioning as a binary classifier, which outputs the likelihood two given poses match.  We govern training with a binary cross entropy loss providing our network only with the supervision of whether the pose pair is a match. See Figure \ref{fig:tokens} for more details.

\subsection{Improved Multi-Frame Pose Estimation} 
\label{estimation}

We now describe how we improve keypoint estimation. Top-down methods suffer from two primary classes of errors from the object detector: 1. Missed bounding boxes 2. Imperfect bounding boxes. We use the box detections from adjacent timesteps in addition to the one in the current timestep to make pose predictions, thereby combating these issues. This is based on the intuition that the spatial location of each person does not change dramatically from frame to frame when the frame rate is relatively high, typical in most modern datasets and cameras. Thus, pasting a bounding box for the $ith$ person in frame, $\mathcal{F}^{t-1}$, $p^{t-1,i}$, in its same spatial location in frame $\mathcal{F}^{t}$ is a good approximation of the true bounding box for person $p^{t,i}$. Bounding boxes are enlarged by a small factor to account for changes in spatial location from frame to frame. Previous approaches, such as ~\cite{xiao2018simple}, use standard non-maximal suppression (NMS) to choose which of these boxes to input into the estimator. Though this addresses the 1st issue of missed boxes, it does not fully address the second issue. NMS relies on the confidence score of the boxes. We make pose predictions for the box in the current frame and temporally adjacent boxes. Then we use object-keypoint similarity (OKS) to determine which of the poses should be kept. This is more accurate than using NMS because we use the confidence scores of the keypoints, not the bounding boxes. The steps of TOKS are enumerated below:

\begin{algorithm}
\caption{Temporal OKS}
\begin{algorithmic} 
\STATE \textbf{Input:} $p^{t-1}, p^t, \mathcal{F}^t$
\STATE 1. Retrieve bounding box, $B$, enclosing $p^{t-1}$, and dilate by a factor, $\alpha$
\STATE 2. Estimate a new pose, $p'^t$, in $\mathcal{F}^t$ from $B$
\STATE 3. Use OKS to determine which pose to keep, $p^* = OKS(p'^t, p^t)$
\STATE \textbf{Output:} $p^*$
\end{algorithmic}
\label{algo:toks}
\end{algorithm}

%% file: sections/results.tex
\section{Experiments}

\subsection{The PoseTrack Dataset}
The \textbf{PoseTrack 2017} training, validation, and test sets consist of 250, 50, and 208 videos, respectively. Annotations for the test set are held out. We evaluate on the PoseTrack 17 Test set because the PoseTrack 18 Test set has yet to be released. We use the official evaluation server on the test set, which can be submitted to up to 4 times. ~\cite{PoseTrack, PoseTrack2017Leaderboard} We conduct the rest of comparisons on the \textbf{PoseTrack ECCV 2018 Challenge Validation Set}, a superset of PoseTrack 17 with 550 training, 74 validation, and 375 test videos~\cite{PoseTrackECCVChallenge}.

\textbf{Metrics} Per-joint Average Precision (AP) is used to evaluate keypoint estimation based on the formulation in ~\cite{MPII}. Multi-Object Tracking Accuracy (MOTA~\cite{bernardin2008evaluating},~\cite{MOTA}) scores tracking and penalizes False Negatives (FN), False Positives (FP), and ID Switches (IDSW); its formulation for the $ith$ keypoint is given below, where $t$ is the current timestep in the video. Our final MOTA is the average of all keypoints $k^i \in \mathcal{K}$:
\[
 1 - \frac{\sum_t{(FN^i_t + FP^i_t + IDSW^i_t)}}{\sum_tGT^i_t}
\]
Our approach assigns track ids and estimates keypoints independently of one another. This is also true of competing methods with MOTA scores closest to ours. In light of this, we use the same keypoint estimations to compare our Pose Entailment based tracking assignment to competing methods in \ref{sec:track_compare}. This makes the IDSW the only component of the MOTA metric that changes, and we calculate $\%IDSW^i = \sum_t IDSW^i_t  / \sum_t GT^i_t$.  In \ref{sec:tOKS_compare}, we compare our estimation method to others without evaluating tracking. Finally, in \ref{pipeline_compare}, we compare our entire tracking pipeline to other pipelines.

\subsection{Improving Tracking with Pose Entailment} \label{sec:track_compare}
\begin{figure*}[h]
\centering
\renewcommand{\arraystretch}{1.25}
\begin{adjustbox}{width=1.0\textwidth}
    \begin{tabular}{r l c c c c g}
        \multicolumn{6}{c}{\large{~~~~~~~~~~~~~~~~~PoseTrack 2018 ECCV Challenge Val Set}} \\
        \toprule
        No. & Method &  Extra Data & AP$^T$ & AP & FPS & MOTA  \\
        \midrule
        1. & \textbf{KeyTrack (ours)} & \xmark & 74.3 & \textbf{81.6} & 1.0 & \textbf{66.6} \\
        2. & MIPAL \cite{hwang2019pose} & \xmark & \bf 74.6 & - & - & 65.7 \\
        3. & LightTrack (offline) \cite{ning2019lighttrack} & \xmark & 71.2 & 77.3 & E & 64.9 \\
        4. & LightTrack (online) \cite{ning2019lighttrack} & \xmark & 72.4 & 77.2 & 0.7 & 64.6 \\
        5. & Miracle \cite{yu2018multi} & \cmark & - & 80.9 & E & 64.0 \\
        6. & OpenSVAI \cite{Ning_2019} & \xmark & 69.7 & 76.3 & - & 62.4 \\
        7. & STAF \cite{raaj2019efficient} & \cmark & 70.4 & - & \textbf{3} & 60.9 \\
        8.  & MDPN \cite{Guo_2019} & \cmark & 71.7 & 75.0 & E & 50.6 \\
        \bottomrule
    \end{tabular}
    \quad
    \quad
    \begin{tabular}{r l c c c g}
        \multicolumn{5}{c}{\large{~~~~~~~~~~~~~~~PoseTrack 2017 Test Set Leaderboard}} \\
        \toprule
        No. & Method & Extra Data & AP$^T$ & FPS & MOTA  \\
        \midrule
        1. & \textbf{KeyTrack (ours)} & \xmark & 74.0 & 1.0 & \textbf{61.2} \\
        2. & POINet \cite{Ruan:2019:PPO:3343031.3350984} & \xmark & 72.5 & - & 58.4 \\
        3. & LightTrack \cite{ning2019lighttrack} & \xmark & 66.7 & E & 58.0 \\
        4. & HRNet \cite{HRNet} & \xmark & \bf 75.0 & 0.2 & 57.9 \\
        5. & FlowTrack \cite{xiao2018simple} & \xmark & 74.6 & 0.2 & 57.8 \\
        6. & MIPAL \cite{hwang2019pose} & \xmark & 68.8 & - & 54.5 \\
        7. & STAF \cite{raaj2019efficient} & \cmark & 70.3 & \textbf{2} & 53.8 \\
        8. & JointFlow \cite{doering2018joint} & \xmark & 63.6 & 0.2 & 53.1 \\
        \bottomrule
    \end{tabular}
\end{adjustbox}
\caption{Top scores on the PoseTrack leaderboards. E indicates an ensemble of detectors is used, and results in the method being offline. A check indicates external training data is used beyond COCO and PoseTrack. A ``-" indicates the information has not been made publicly available. FPS calculations for JointFlow and FlowTrack are taken from \cite{zhang2019fastpose}. HRNet FPS is approximated from FlowTrack since the methods are very similar. The AP column has the best AP score. AP$^T$ is the AP score after tracking post-processing.}
\label{Tab:leaderboard}
\end{figure*}
\begin{figure*}
    \centering
    \includegraphics[width=0.24\textwidth,trim={8cm 5cm 8cm 5cm},clip]{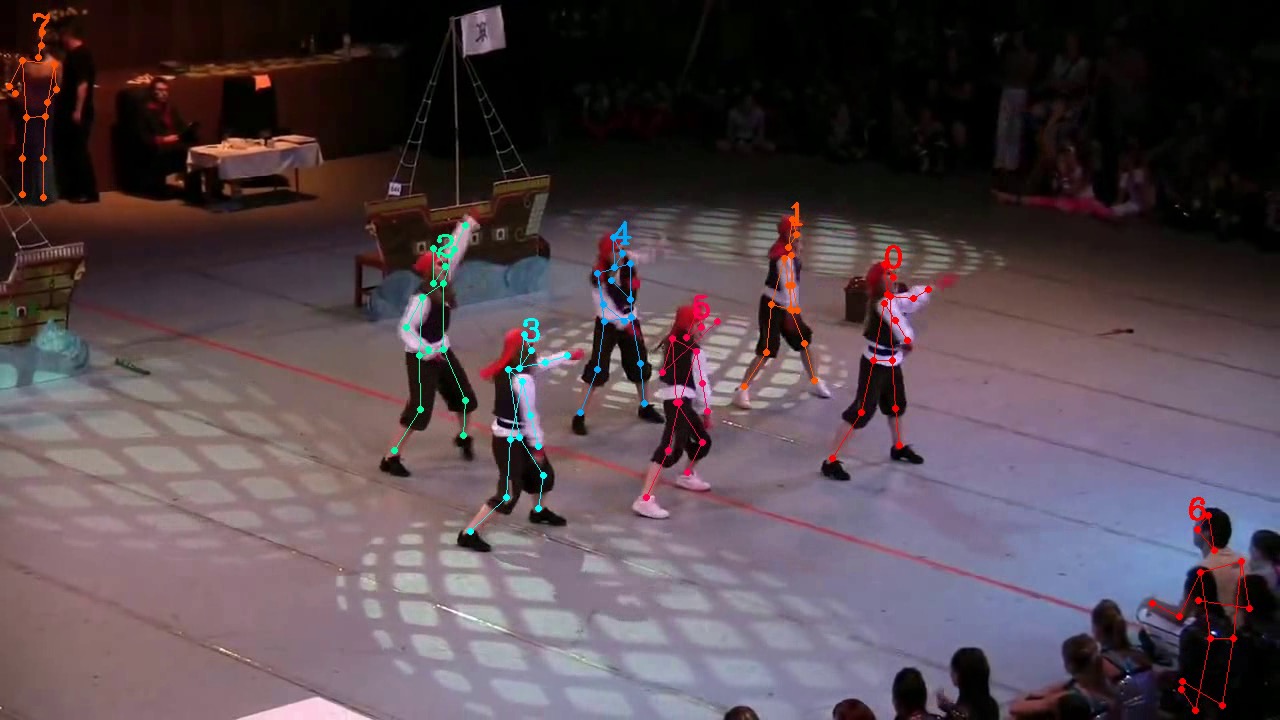}
    \includegraphics[width=0.24\textwidth,trim={8cm 5cm 8cm 5cm},clip]{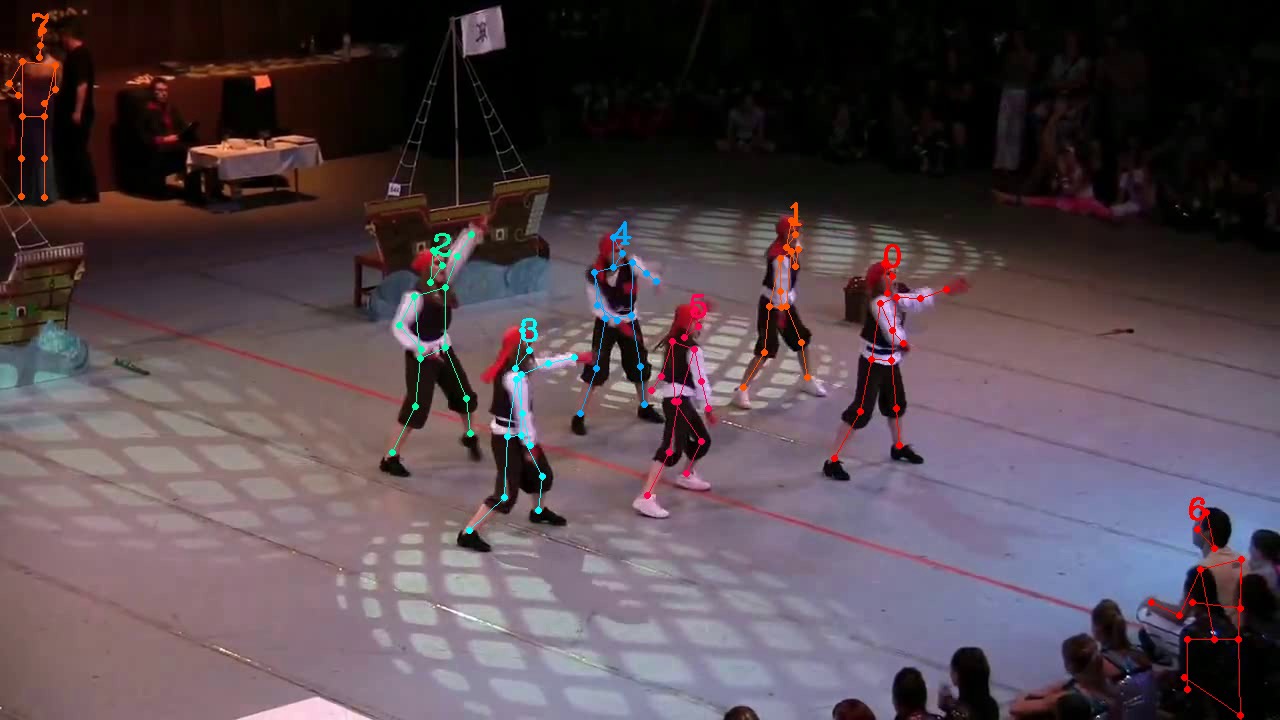}
    \includegraphics[width=0.24\textwidth,trim={8cm 5cm 8cm 5cm},clip]{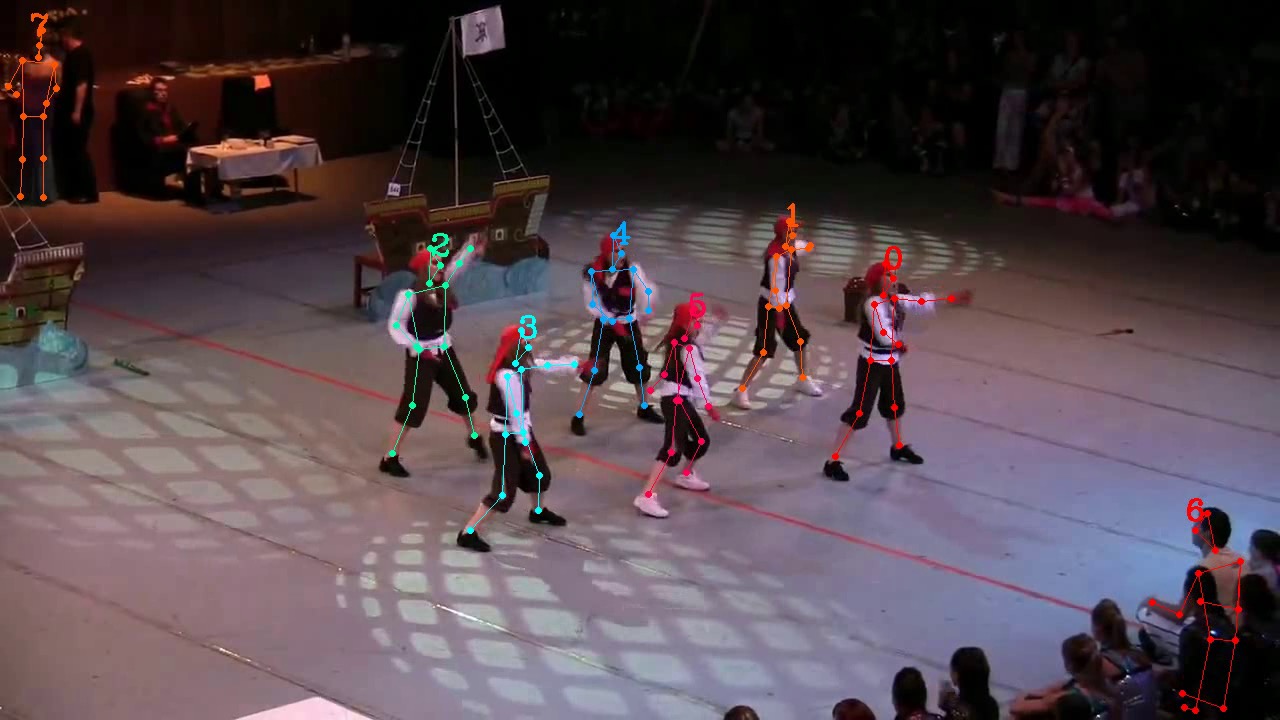}
    \includegraphics[width=0.24\textwidth,trim={8cm 5cm 8cm 5cm},clip]{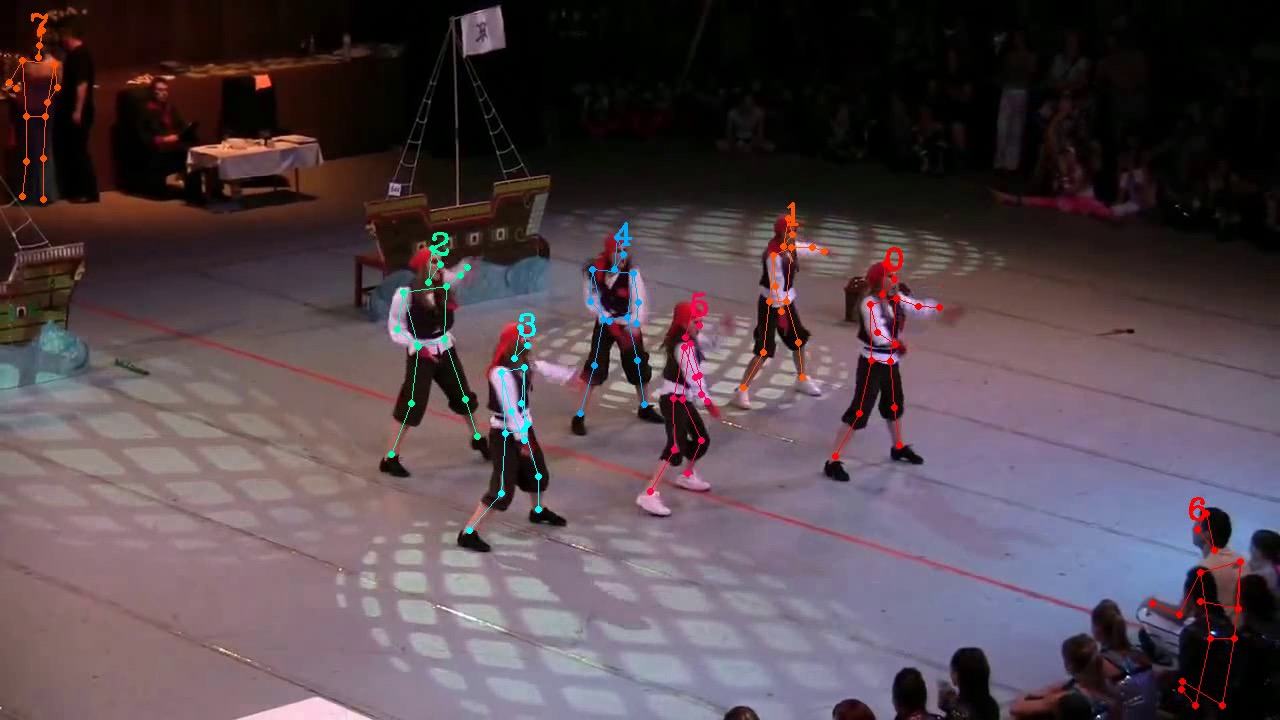}
    % new video
    \\
    \vspace{0.1cm}
    \includegraphics[width=0.24\textwidth]{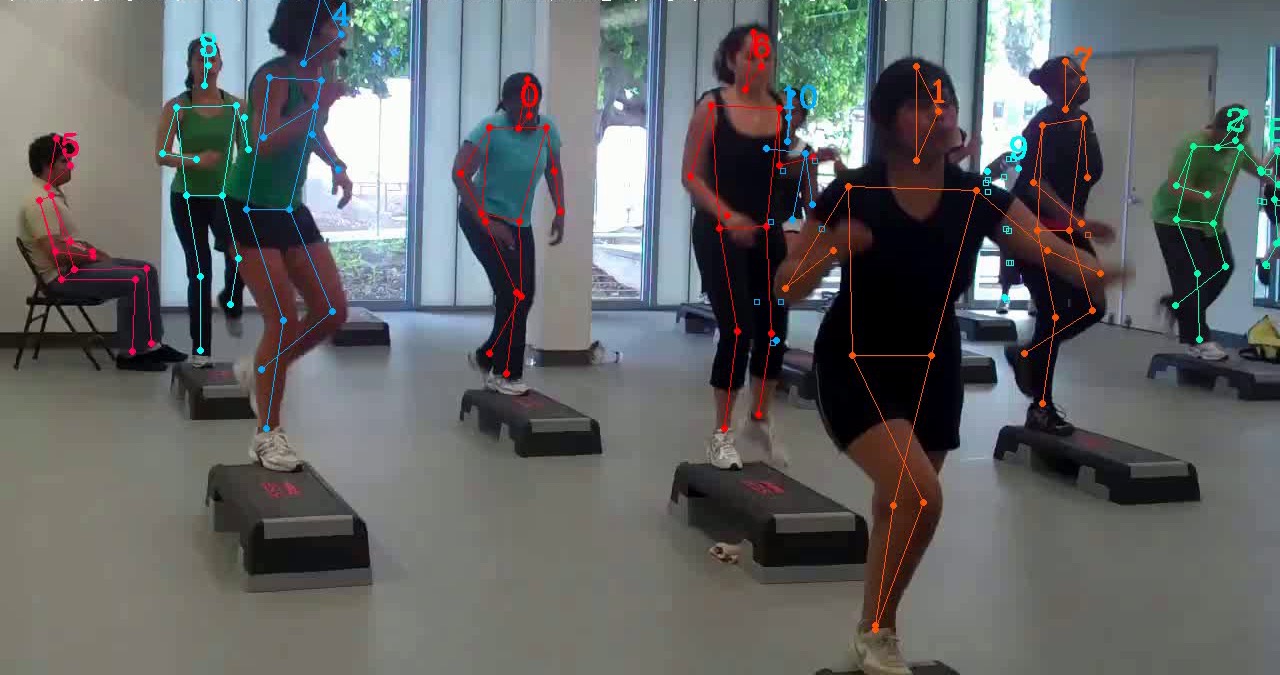}
    \includegraphics[width=0.24\textwidth]{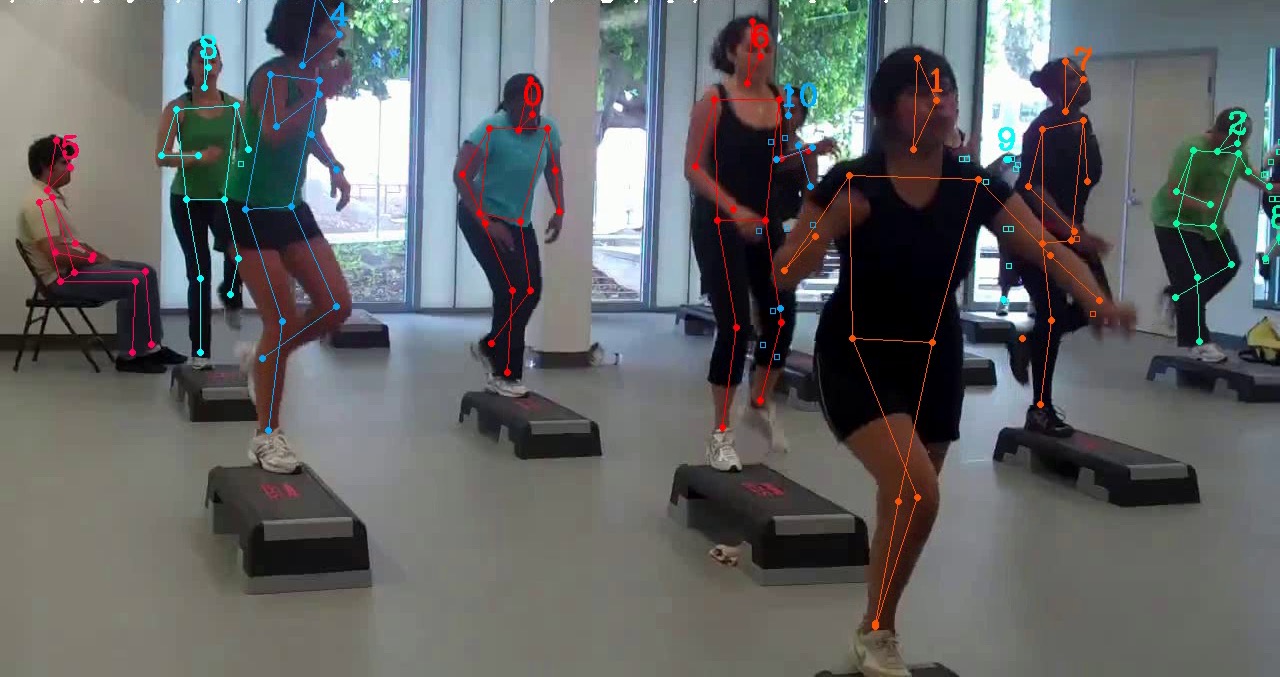}
    \includegraphics[width=0.24\textwidth]{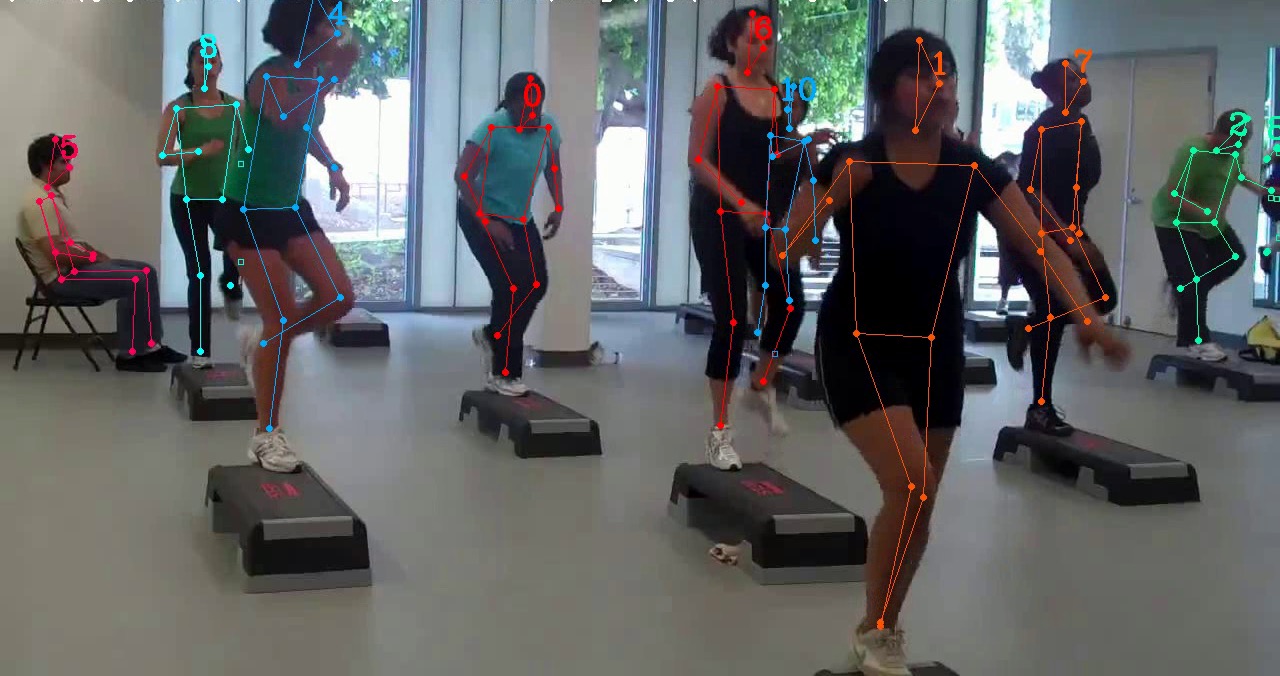}
    \includegraphics[width=0.24\textwidth]{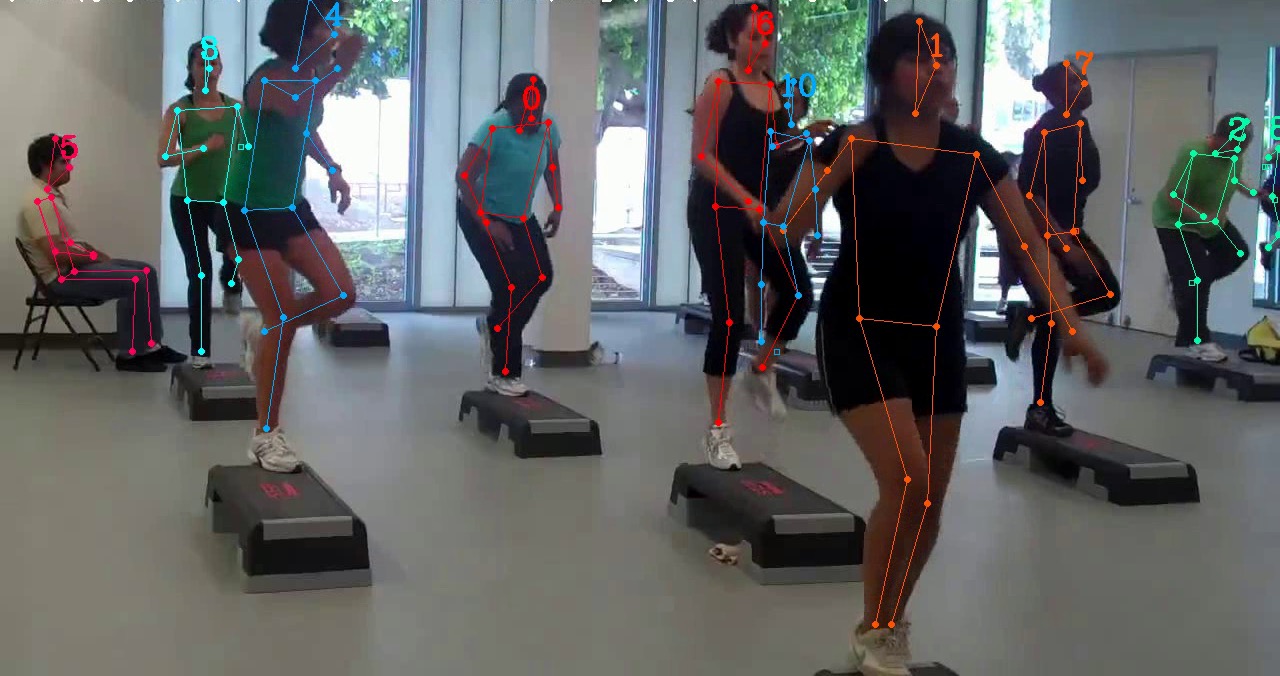}
    \label{fig:video_seq}
    \caption{Qualitative results of \trackname, on the PoseTrack 18 Validation Set (top row) and PoseTrack 17 Test Set (bottom row).}
\end{figure*}{}
We compare with the optical flow tracking method~\cite{xiao2018simple}, and the Graph Convolutional Network~\cite{ning2019lighttrack} (GCN) as shown in Figure~\ref{tab:idsw}. We do not compare with IoU because our other baselines, GCN and optical flow \cite{ning2019lighttrack}, \cite{xiao2018simple} have shown to outperform it, nor do we compare to the network from \cite{Ruan:2019:PPO:3343031.3350984} because it is trained in an end-to-end fashion.  We follow the method in \cite{xiao2018simple} for Optical Flow and use the pre-trained GCN provided by \cite{ning2019lighttrack}. IDSW is calculated with three sets of keypoints. Regardless of the keypoint AP, we find that \trackname\textquotesingle s Pose Entailment maintains a consistent improvement over other methods. We incur approximately half as many IDSW as the GCN and 30\% less than Optical Flow.

Our improvement over GCN stems from the fact that it relies only on keypoint spatial locations. By using additional information beyond the spatial location of each keypoint, our model can make better inferences about the temporal relationship of poses. The optical flow CNNs are not specific to pose tracking and require manual tuning. For example, to scale the CNN's raw output, which is normalized from -1 to 1, to pixel flow offsets, a universal constant, given by the author of the original optical flow network (not \cite{xiao2018simple}), must be applied. However, we found that this constant did not produce good results and required manual adjustment. In contrast, our learned method requires no tuning during inference. 

\subsection{Improving Detection with TOKS} \label{sec:tOKS_compare}

\begin{table}[t]
    \renewcommand{\arraystretch}{1.1}
    \resizebox{\linewidth}{!}{
    \begin{tabular}{l | c c c c c c c c}
        \toprule
        \multirow{2}{*}{Detection Method} & \multicolumn{8}{c}{AP} \\
        & Head & Shou & Elb & Wri & Hip & Knee & Ankl & Total \\
        \midrule
        GT & 90.2 & 91.4 & 88.7 & 83.6 & 81.4 & 86.1 & 83.7 & 86.7 \\
        \hdashline
        Det. & 68.8 & 72.8 & 73.1 & 68.4 & 68.0 & 72.4 & 69.8 & 70.4 \\
        Det. + Box Prop. & 79.3 & 82.0 & 80.8 & 75.6 & 72.4 & 76.5 & 72.4 & 77.1 \\
        Det. + TOKS@0.3 & 83.6 & 86.6 & 84.9 & 78.9 & 76.4 & 80.2 & 76.2 & 81.1 \\
        \bf Det. + TOKS@0.35 (ours) & \bf 84.1 & \bf 87.2 & \bf 85.3 & \bf 79.2 & \bf 77.1 & 80.6 & \bf 76.5 & \bf 81.6 \\
        Det. + TOKS@0.5 & 83.9 & 87.2 & 85.2 & 79.1 & 77.1 & \bf 80.7 & 76.4 & 81.5 \\
        \bottomrule
    \end{tabular}
    }
    \caption{Per-joint AP when the pose estimator is conditioned on different boxes. GT indicates ground truth boxes are used, and serves as an upper bound for accuracy. Det. indicates a detector was used to estimate boxes. @OKS* is the OKS threshold used.}
    \label{Tab:detection}
\end{table}

Table \ref{Tab:detection} shows offers a greater improvement in keypoint detection quality than other methods. In the absence of bounding box improvement, the AP performance is  6.6\% lower, highlighting the issue of False Negatives. The further improvement from TOKS emphasizes the usefulness of estimating every pose. By using NMS, bounding box propagation methods miss the opportunity to use the confidence scores of the keypoints, which lead to better pose selection.

\subsection{Tracking Pipeline Comparison to the SOTA} \label{pipeline_compare}

Now that we have analyzed the benefits of Pose Entailment and TOKS, we put them together and compare to other approaches. Figure \ref{Tab:leaderboard} shows that we achieve the highest MOTA score. We improve over the original HRNet paper by 3.3 MOTA points on the Test set. \cite{hwang2019pose}, nearest our score on the 2018 Validation set, is much further away on the 2017 Test set. Additionally, our FPS is improved over all methods with similar MOTA scores, with many methods being offline due to their use of ensembles. (Frames per second (FPS) is calculated by diving the number of frames in the dataset by the runtime of the  approach.) Moreover, our method outperforms all others in terms of AP, showing the benefits of TOKS. AP$^T$ is also reported, which is the AP score after tracking post-processing has been applied. This post-processing is beneficial to the MOTA score, but lowers AP. See \ref{sec:postprocessing} for more details on this post-processing. As we have the highest AP, but not the highest AP$^T$ it appears the effect of tracking post-processing varies from paper to paper. Only AP$^T$ is given on the test set because each paper is given 4 submissions, so these are used to optimize MOTA, rather than AP. 

\paragraph{Efficiency:} Our tracking approach is efficient, not reliant on optical flow or RGB data. When processing an image at our optimal resolution, 24x18, we reduce the GFLOPS required by optical flow, which processes images at full size, from 52.7 to 0.1. \cite{ning2019lighttrack}'s GCN does not capture higher-order interactions over keypoints and can be more efficient than our network with local convolutions. However, this translates to a $\sim$1ms improvement in GPU runtime. In fact, with other optimizations, our tracking pipeline demonstrates a 30\% improvement in end-to-end runtime over \cite{ning2019lighttrack}, shown in \ref{pipeline_compare}. We have the fastest FPS of Top-down approaches. Bottom-up models such as STAF, are more efficient but have poor accuracy. Also, we do not rely on optical flow to improve bounding box propagation as \cite{xiao2018simple, HRNet} do, instead we use TOKS. This contributes to our 5x FPS improvement over \cite{xiao2018simple, HRNet}. Further details on the parameters and FLOPS of the GCN, Optical Flow Network, and our Transformer Matching Network are in Table \ref{tab:flops}.

%% file: sections/analysis.tex
\section{Analysis}
\label{sec:analysis}

\subsection{Tracking Pipeline} \label{sec:pipeline_study}

\begin{figure*}[t!]
    \begin{adjustbox}{width=0.45\textwidth}
    \begin{tabular}{c c c c | c c}
        \toprule
        Num Tx & Hidden Size & Int. Size & Num Heads & Parameters (M) & \% IDSW \\
        \midrule
        2 & 128 & 512 & 4 & 0.40 & 1.0 \\
        4 & 128 & 512 & 4 & 0.43 & \bf 0.8 \\
        6 & 128 & 512 & 4 & 1.26 & 1.1 \\
        \hdashline
        4 & 64 & 256 & 4 & 0.23 & 0.9 \\
        4 & 128 & 512 & 4 & 0.43 & \bf 0.8 \\
        4 & 256 & 1024 & 4 & 3.31 & 1.1 \\
        \hdashline
        4 & 128 & 128 & 4 & 0.43 & \bf 0.8 \\
        4 & 128 & 512 & 4 & 0.86 & 0.8 \\
        \hdashline
         4 & 128 & 128 & 2 & 0.43 & 0.9 \\
         4 & 128 & 128 & 4 & 0.43 & \bf 0.8 \\  
         4 & 128 & 128 & 6 & 0.43 & 0.8 \\  
        \bottomrule
    \end{tabular}
    \end{adjustbox}
    \quad
    \quad
    \raisebox{-1.7cm}{\includegraphics[width=0.5\textwidth]{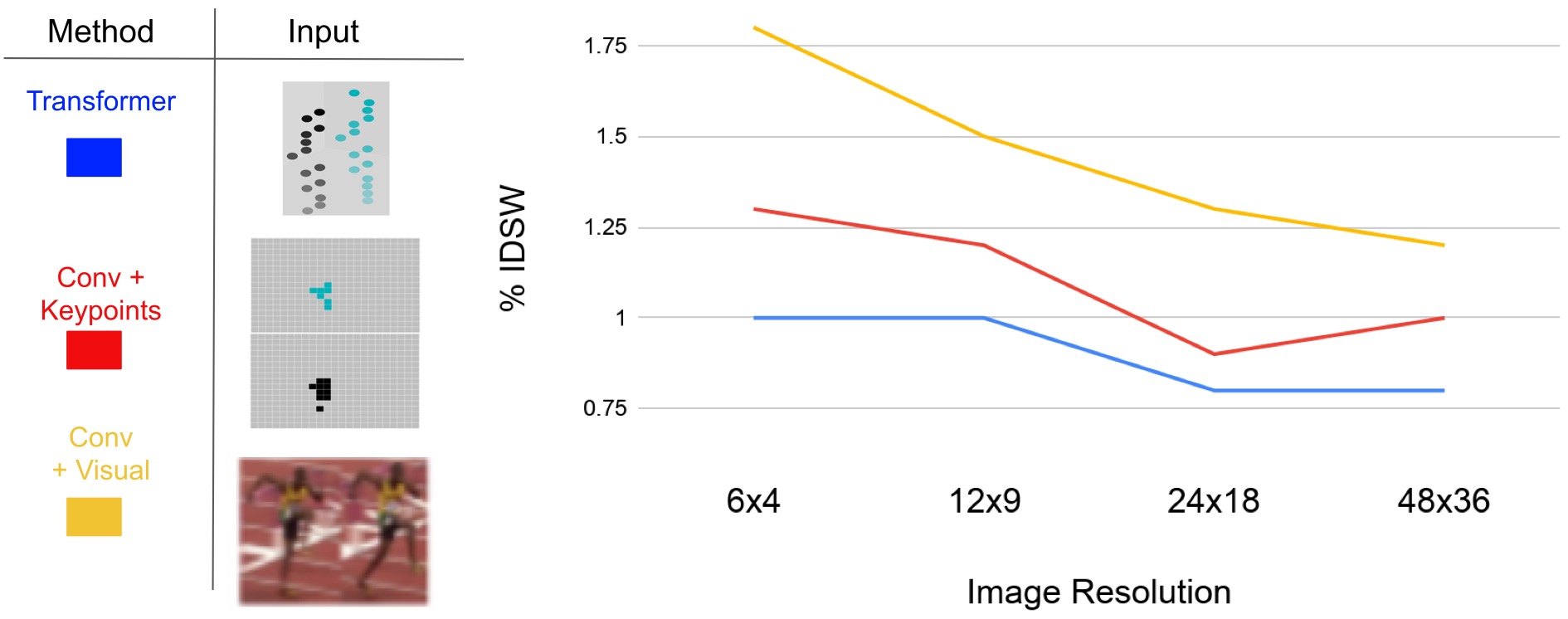}}
    \caption{\textbf{Left:} Transformer network hyper-parameters are varied. \textbf{Right:} A plot of IDSW rate vs. image resolution. The table on the left shows the input to each method, the conv+visual input is blurry because images are downsampled.}
    \label{fig:analysis}
\end{figure*}

\textbf{Varying Tokenization Schemes and Transformer Hyper-parameters} We examine the benefits of each embedding. As evident in Table \ref{Tab:embeddings}, \emph{Segment} embeddings are crucial because they enable the network to distinguish between the Poses being matched. \emph{Token} embeddings give the network information about the orientation of a pose and help it interpret keypoints which are in close spatial proximity; i.e. keypoints that have the same or similar position embedding. We also train a model that uses the relative keypoint distance from the pose center rather than the absolute distance of the keypoint in the entire image. We find that match accuracy deteriorates with this embedding. This is likely because many people perform the same activity, such as running, in the PoseTrack dataset, leading to them having nearly identical relative pose positions. We vary the number of transformer blocks, the hidden size in the transformer block, and number of heads in Table \ref{fig:analysis}. Decreasing the number of transformer blocks, the hidden size, and attention heads hurts performance.

\begin{table}[t]
    \renewcommand{\arraystretch}{1.4}
    \resizebox{\linewidth}{!}{
    \begin{tabular}{c c c c c}
        \toprule
        Abs. Position & Type & Segment & Rel. Position & Match \% Accuracy \\
        \midrule
        \cmark & \cmark & \xmark & \xmark & 72.6 \\
        \cmark & \xmark & \cmark & \xmark & 90.0 \\
        \cmark & \cmark & \cmark & \xmark & \bf{93.2 (ours)} \\
        \xmark & \cmark & \cmark & \cmark & 91.3 \\
        \cmark & \cmark & \cmark & \cmark & 92.0 \\
        \bottomrule
    \end{tabular}
    }
    \caption{Match accuracies for various embedding schemes.}
    \label{Tab:embeddings}
\end{table}

\begin{figure}
    \centering
    \includegraphics[width=0.4\textwidth]{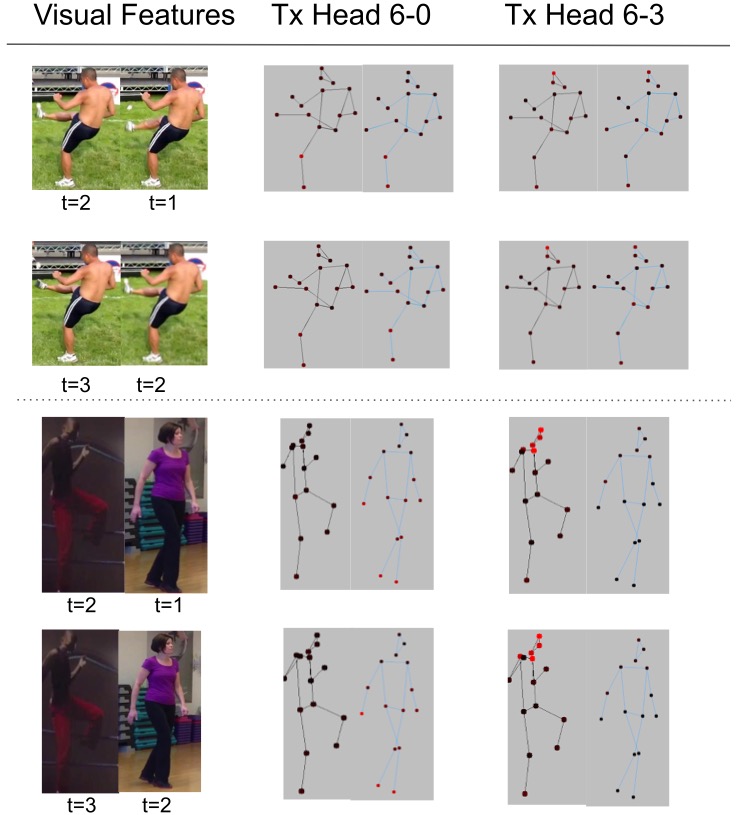}
    \caption{Attention heatmaps from two of our network's attention heads are shown. These are the 0th, and 3rd heads from our final transformer. The two pairs above the dotted line are a matching pair, while the pair below the dotted line are not (and are also from separate videos). $t$ is the frame timestep.}
    \label{fig:heatmaps}
\end{figure}{}

\textbf{Number of Timesteps and Other Factors} We find that reducing the number of timesteps adversely effects the MOTA score. It drops up to 0.3 points when using only a single timestep because we are less robust to detection errors. Also, in replacement of our greedy algorithm, we experimented with the Hungarian algorithm used in \cite{girdhar2018detect}. This algorithm is effective with ground truth information, but is not accurate when using detected poses.

\subsection{Comparing Self-Attention to Convolutions} \label{sec:self-attention} 
We compare transformers and CNNs by replacing our Transformer Matching Network with two separate convolution-based networks. One takes visual features from bounding box pose pairs as input while the other takes only keypoints as input, where each unique keypoint is colored via a linear interpolation, a visual version of our \emph{Type} tokens. Both approaches use identical CNNs, sharing an architecture inspired by VGG \cite{simonyan2014deep}, and have approximately 4x more parameters than our transformer-based model because this was required for stable training. See \ref{sec:details} for details.

Transformers outperform CNNs for the tracking task, as shown in Figure \ref{fig:analysis}. However, we find two areas where CNNs can be competitive. First, at higher resolutions, transformers often need a large number of parameters to match CNN's performance. In NLP, when using large vocabularies, a similar behavior is observed where transformers need multiple layers to achieve good performance. Second, we also find  that convolutions optimize more quickly than the transformers, reaching their lowest number of ID Switches within the first 2 epochs of training. Intuitively, CNNs are more easily able to take advantage of spatial proximity. The transformers receive spatial information via the position embeddings, which are 1D linear projections of 2D locations. This can be improved by using positional embedding schemes that better preserve spatial information~\cite{girdhar2019video}. 

In summary, CNNs are accurate at high resolutions given its useful properties such as translation invariance and location invariance. However, there is an extra computational cost of using them. The extra information, beyond the spatial location of keypoints, included in our keypoint embeddings, coupled with the transformer's ability to model higher-order interactions allows it to function surprisingly well at very low resolutions. Thus, the advantage of CNNs is diminished and our transformer-based network outperforms them in the low resolution case. 

\subsection{Visualizing Attention Heatmaps} \label{sec:heatmap} We visualize our network's attention heatmaps in Fig. \ref{fig:heatmaps}. When our network classifies a pair as non-matching, its attention is heavily placed on one of the poses over the other. Also, we find it interesting that one of our attention heads primarily places its attention on keypoints near the person's head. This specialization suggests different attention heads are attuned to specific keypoint motion cues.

%% file: sections/conclusion.tex
\section{Conclusion}

In this paper, we present an efficient Multi-person Pose Tracking method. 
Our proposed Pose Entailment method 
achieves SOTA performance on the PoseTrack datasets by using keypoint information
in the tracking step without the need of optical flow or CNNs. \trackname\ also benefits from improved keypoint estimates using our temporal refinement method that outperforms commonly
used bounding box propagation methods. Finally, we also demonstrate how to tokenize and embed human pose information in the transformer architecture that can be re-used for other tasks such as
pose-based action recognition.